\pdfoutput=1
\documentclass[11pt]{article}
\usepackage[pageonlyfooter]{evolvingmachines}
\usepackage{algorithm}

\usepackage{algorithmic}
\usepackage{makecell}
\usepackage{multirow}
\usepackage{subcaption}
\usepackage{pifont}
\usepackage{fontawesome5}
\usepackage{adjustbox}
\usepackage{placeins}
\usepackage[nameinlink,capitalize,noabbrev]{cleveref}
\let\cite\citep
\newcommand{\fitwidth}[1]{\begin{adjustbox}{max width=\linewidth}#1\end{adjustbox}}

\definecolor{ForestGreen}{RGB}{34,139,34}
\newcommand{\cmark}{\ding{51}}

\newcommand{\gaincell}[2]{\makecell{\textbf{#1}\\{\fontsize{7}{8}\selectfont\color{red}(#2)}}}
\newcommand{\dropcell}[2]{\makecell{#1\\{\fontsize{7}{8}\selectfont\color{ForestGreen}(#2)}}}
\newcommand{\samecell}[1]{\makecell{#1\\{\fontsize{7}{8}\selectfont\color{black!55}(+0.00)}}}
\newcommand{\thinkmodel}[1]{\makecell[l]{\textbf{Think3D}\\(#1)}}

\providecommand{\Description}[2][]{\ignorespaces}
\DeclareUnicodeCharacter{2192}{\ensuremath{\rightarrow}}

\emltitlesize{19}{23}
\title{Think3D: Thinking with Space for Spatial Reasoning}

\author{\texorpdfstring{%
\makebox[\linewidth][s]{%
Zaibin~Zhang\textsuperscript{1,*,\textdagger},\hfill Yuhan~Wu\textsuperscript{1,*},\hfill
Lianjie~Jia\textsuperscript{1,*},\hfill Yifan~Wang\textsuperscript{1},\hfill
Zhongbo~Zhang\textsuperscript{1},\hfill Yijiang~Li\textsuperscript{2,\textdagger}}\\
\makebox[\linewidth][s]{%
Binghao~Ran\textsuperscript{1},\hfill Fuxi~Zhang\textsuperscript{1},\hfill
Zhuohan~Sun\textsuperscript{1},\hfill Yizhuang~Peng\textsuperscript{4},\hfill
Zhenfei~Yin\textsuperscript{3},\hfill Lijun~Wang\textsuperscript{1},\hfill
Huchuan~Lu\textsuperscript{1}}%
}{Zaibin Zhang; Yuhan Wu; Lianjie Jia; Yifan Wang; Zhongbo Zhang; Yijiang Li; Binghao Ran; Fuxi Zhang; Zhuohan Sun; Yizhuang Peng; Zhenfei Yin; Lijun Wang; Huchuan Lu}}
\emlaffiliation{%
\makebox[\linewidth][c]{%
\textsuperscript{1}Dalian University of Technology\hspace{5mm}%
\textsuperscript{2}University of California San Diego}\\
\makebox[\linewidth][c]{%
\textsuperscript{3}University of Oxford\hspace{5mm}%
\textsuperscript{4}Independent Researcher\hspace{5mm}%
\textsuperscript{5}Evolving Machines Lab}\\[1pt]
\makebox[\linewidth][c]{%
\footnotesize\textsuperscript{*}Equal contribution (joint first authors).
\hspace{5mm}\textsuperscript{\textdagger}Project leader.}}
\emlcorrespondence{%
\makebox[\linewidth][c]{%
\href{https://github.com/zhangzaibin/spagent}{\faGithub\enspace Code, Models \& Data}%
\hspace{6mm}%
\href{mailto:dlutzzb@gmail.com}{\faEnvelope\enspace dlutzzb@gmail.com}}\\
\makebox[\linewidth][c]{%
\href{mailto:tracy1252684562@mail.dlut.edu.cn}{tracy1252684562@mail.dlut.edu.cn},\quad
\href{mailto:jialianjie@mail.dlut.edu.cn}{jialianjie@mail.dlut.edu.cn},\quad
\href{mailto:ljwang@dlut.edu.cn}{ljwang@dlut.edu.cn}}}
\emlshorttitle{Think3D: Thinking with Space for Spatial Reasoning}
\emlkeywords{Spatial reasoning; vision-language agents; 3D tool use; active perception; reinforcement learning}
\date{}

\emlheader{%
  \begin{minipage}[c]{10mm}
    \centering\includegraphics[height=10mm]{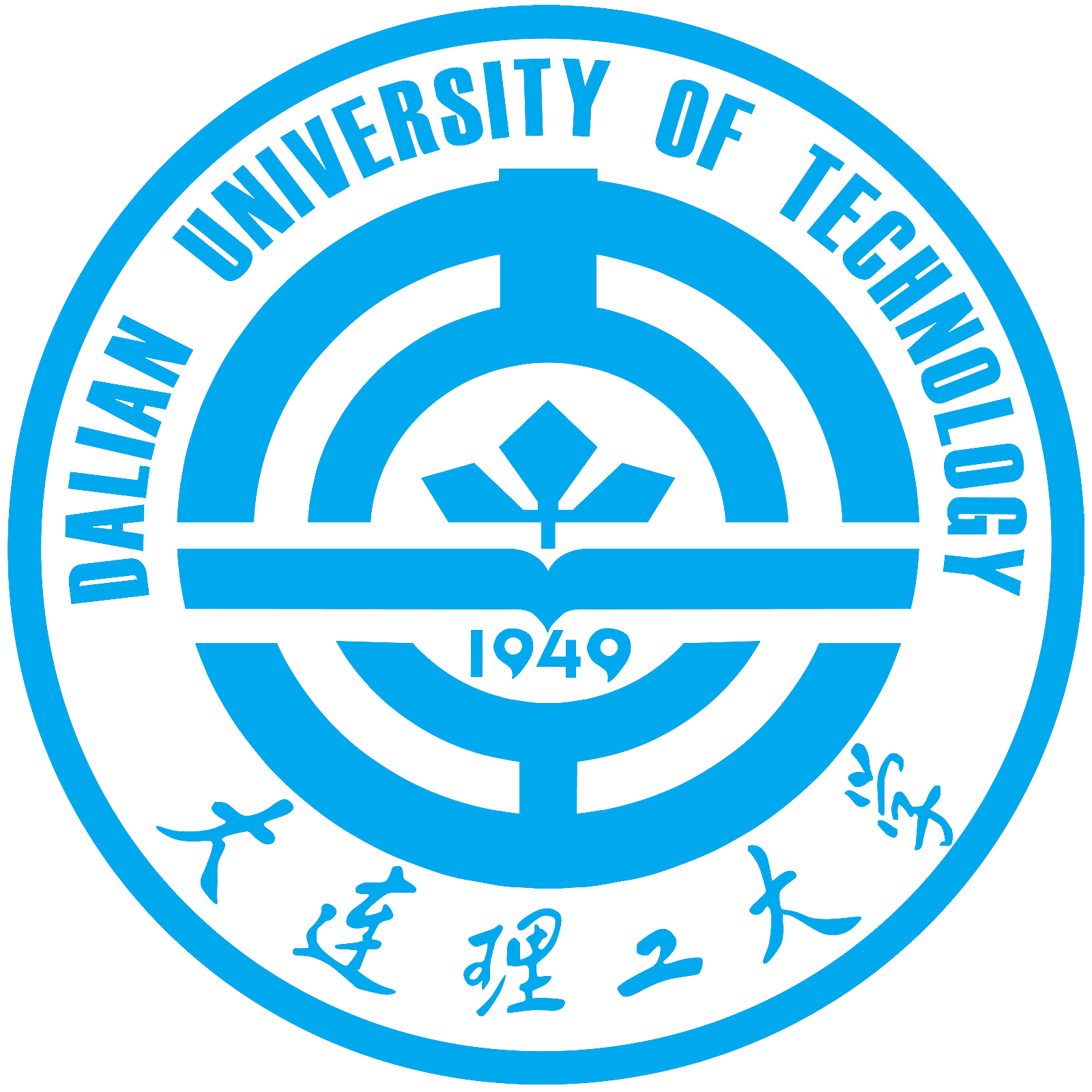}%
  \end{minipage}\hspace{5mm}%
  \begin{minipage}[c]{34mm}
    \centering
    \includegraphics[width=34mm,trim={346bp 644bp 346bp 644bp},clip]{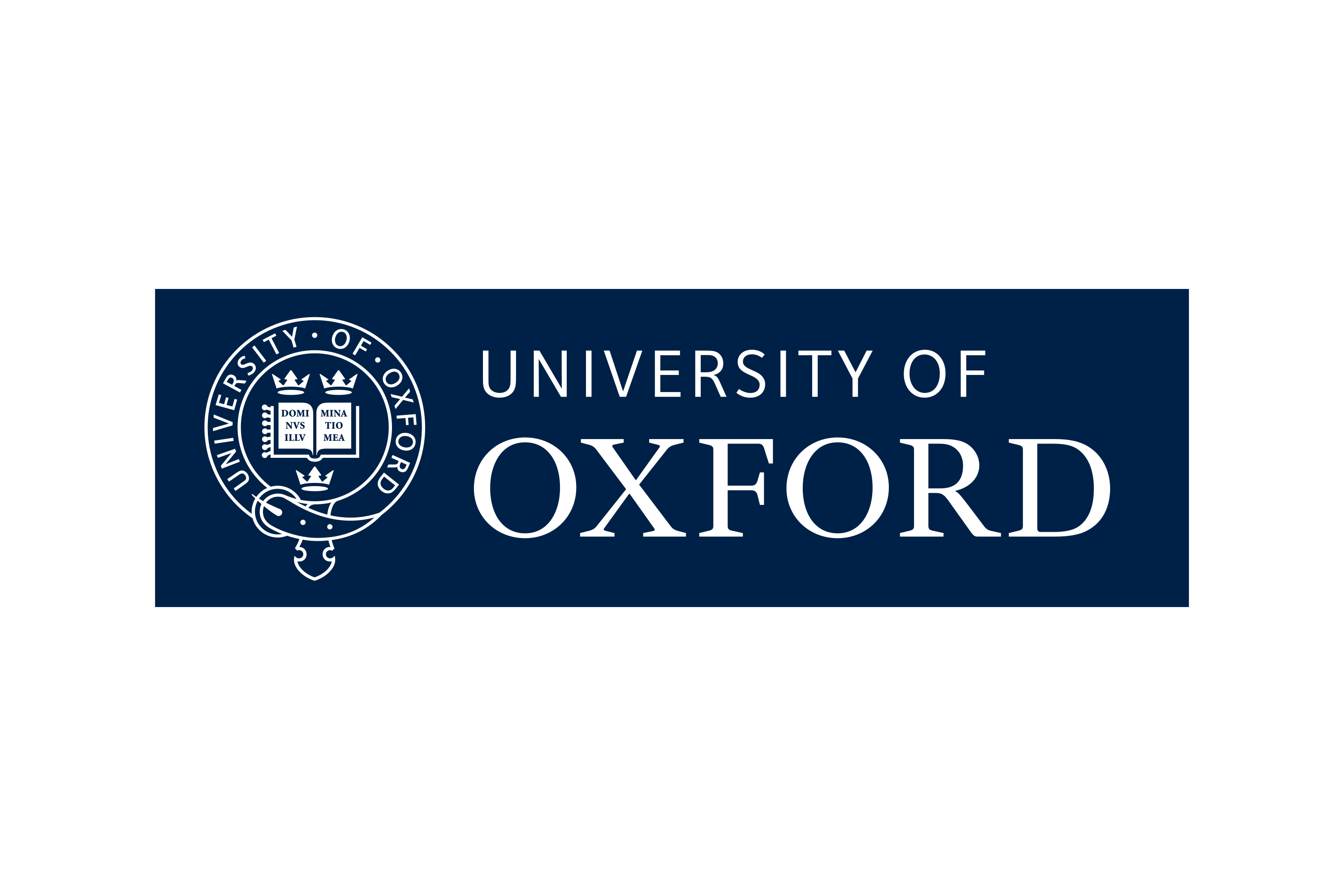}%
  \end{minipage}\hspace{5mm}%
  \begin{minipage}[c]{24mm}
    \centering\includegraphics[width=24mm]{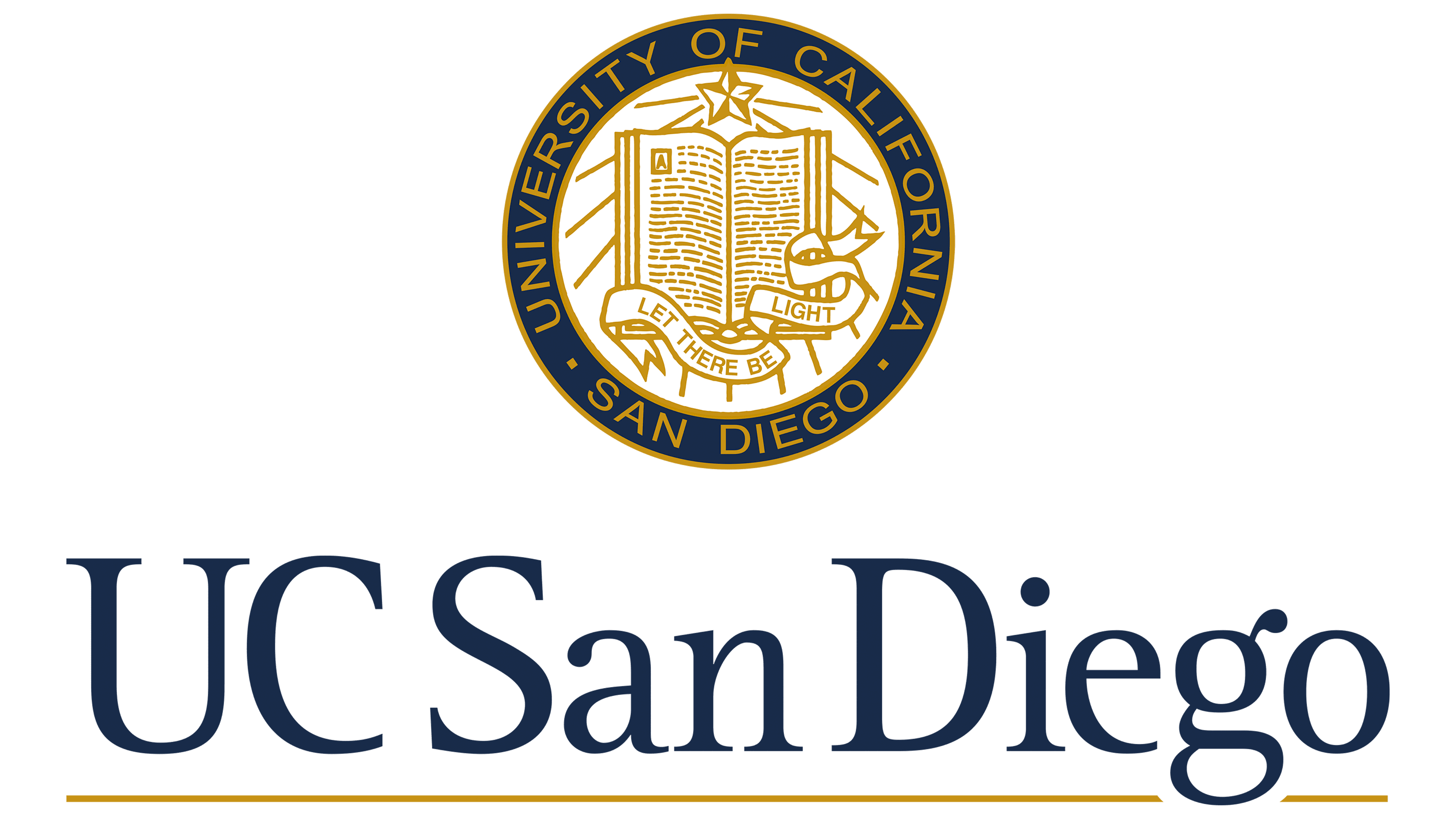}%
  \end{minipage}\hspace{5mm}%
  \begin{minipage}[c]{32mm}
    \centering
    \resizebox{32mm}{!}{%
      \begin{tabular}[c]{@{}c@{\hspace{3.5mm}}l@{}}
        \begin{tabular}[c]{@{}c@{}}\EMLMark[12mm]\end{tabular} & \EMLWordmark
      \end{tabular}}%
  \end{minipage}%
}

\begin{document}
\begingroup
\setlength{\parskip}{1.5pt}
\maketitle
\endgroup

\noindent\begin{minipage}{\linewidth}
    \captionsetup{type=figure}
    \centering
    \includegraphics[width=\linewidth,height=80mm,keepaspectratio]{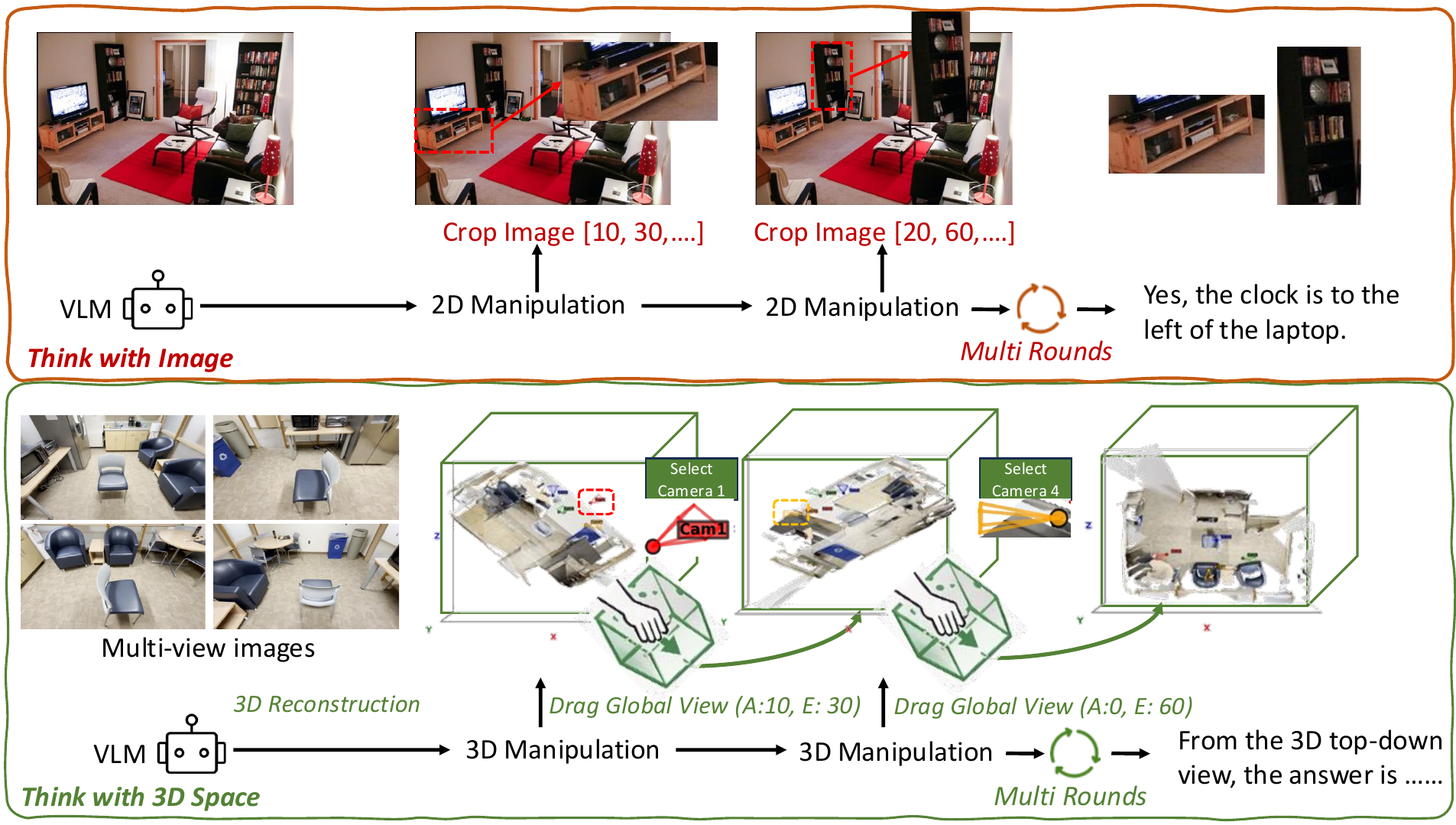}
    \caption{Comparison between prior ``think with image''~\cite{zheng2025deepeyes, su2025thinking} and our ``think with 3D space''. While the former reasons over 2D content by manipulating images, our method operates directly within 3D point cloud space for spatial understanding.}
    \Description{Comparison of image-based visual reasoning and Think3D. The former crops and manipulates two-dimensional images, while Think3D reconstructs a point cloud and requests alternative views to answer a spatial question.}
    \label{fig:teaser}
\end{minipage}\par

\vspace{1mm}
\begin{abstract}
While Vision-Language Models (VLMs) excel at 2D visual understanding, they remain constrained by 2D-centric paradigm that severely limits genuine 3D spatial reasoning. To bridge this gap, we introduce Think3D, a novel framework that equips VLM agents with interactive, 3D chain-of-thought reasoning capabilities. By integrating a suite of 3D manipulation tools, Think3D transforms perception into active spatial exploration, mirroring human geometric reasoning. Think3D consistently improves proprietary models, including GPT-4.1 and Gemini 2.5 Pro, across BLINK Multi-view, MindCube-1K, and VSI-Bench-Tiny. We further propose Think3D-RL to teach smaller open-weight models how to manipulate 3D space effectively. Using only terminal answer and format rewards, without process supervision or handcrafted exploration trajectories, Think3D-RL enables Qwen3-VL-4B to autonomously learn effective 3D exploration strategies. After training, the model exhibits tool-use patterns similar to those of stronger proprietary models, while shifting the effect of 3D tool use on MindCube-1K from a performance drop to a substantial improvement. These results show that active exploration in 3D space provides an effective and general paradigm for improving spatial reasoning in multimodal agents.

\end{abstract}

\clearpage
\section{Introduction}
\label{sec:intro}

Understanding and interacting with the physical world has long been a fundamental objective of vision-language models (VLMs)~\cite{comanici2025gemini,yang2025cambrian,team2026qwen3}. 

Achieving this objective necessitates \emph{spatial intelligence}—the ability to reason about geometry, viewpoint, and spatial relationships~\cite{yang2025thinking, feng2025survey, yu2025far,cai2025scaling}.

Despite remarkable progress in visual understanding, current VLMs remain powerful yet fundamentally \emph{2D analyzers}. Their performance drops sharply on tasks that require spatial reasoning, such as multi-view understanding and route planning. For instance, although recent models achieve near human-level performance on comprehensive benchmarks like MMMU~\cite{yue2024mmmu}, they still lag far behind humans on tasks that demand genuine 3D reasoning~\cite{yang2025thinking, yin2025spatial}.

Existing efforts mainly follow two directions. The first seeks to internalize spatial knowledge by training models on large-scale spatial and embodied datasets~\cite{chen2024spatialvlm,fan2025vlm,ji2025robobrain,team2025gemini,zhou2025robotracer}. Although effective, this approach requires substantial training resources and may affect the general capabilities of the model. The second direction augments VLMs with external tools. Some methods manipulate 2D images to facilitate visual reasoning~\cite{zheng2025deepeyes,su2025thinking}, whereas others integrate 2D and 3D tools through code execution or multi-agent workflows~\cite{han2025tiger,luo2026pySpatial,chen2026geometrically,cho2026spatialclaw,dai2026sagent}. This motivates a simple question: without elaborate agent architectures, can a VLM reason about 3D scenes by directly manipulating the 3D space? In other words, can a VLM think with space?

Recent advances in 3D reconstruction~\cite{wang2025vggt, wang2025pi, keetha2025mapanything} make this possible. These models can estimate camera poses and reconstruct 3D point clouds from videos or multi-view images, providing a geometric foundation for explicit spatial reasoning. Building on this foundation, we propose \textbf{Think3D}---a framework that enables VLMs to actively interact with reconstructed 3D point clouds and reason in a spatial manner through \emph{thinking with 3D space}.

Effective 3D reasoning requires a consistent reference frame. We use estimated camera poses as anchors so the model can interpret rotations and directions consistently, avoiding ambiguous spatial manipulations. With this design, the agent can choose a camera, select a rotation, and decide where to explore next, while switching between a global view (overall geometry) and a local view (fine-grained details) to capture both coarse and fine cues. The process is inherently iterative: the model repeatedly interacts with the reconstructed 3D scene, observes new views, and refines its understanding step by step. Through this iterative reasoning process, Think3D develops a coherent spatial representation, mirroring how humans explore 3D space.

We find that the effectiveness of spatial exploration is correlated with the intrinsic reasoning capability of VLMs.
Frontier models such as GPT-4.1~\cite{achiam2023gpt} and Gemini-2.5-Pro~\cite{comanici2025gemini} tend to generate diverse and semantically meaningful viewpoints, whereas less capable models often drift toward redundant or even misleading camera poses, which ultimately limits spatial understanding.
To narrow this gap, we propose a reinforcement learning approach, \textbf{Think3D-RL}, that enables models to autonomously discover effective exploration policies.
Importantly, Think3D-RL uses terminal answer-accuracy and format rewards, without supervision over how the model should manipulate the 3D scene.
During training, the model conducts multi-round spatial exploration, and the reward signal reinforces trajectories that lead to stronger downstream performance.
Through this reward-driven process, the model progressively learns when and how to interact with the 3D environment, converging to substantially more informative viewpoint manipulation strategies.
As a result, models exhibit increasingly consistent exploration behaviors that more closely match those of frontier VLMs, which leads to substantial improvements across diverse spatial reasoning benchmarks.

We evaluate Think3D on three challenging spatial reasoning benchmarks: BLINK Multi-view~\cite{fu2024blink}, MindCube-1K~\cite{yin2025spatial}, and VSI-Bench-Tiny~\cite{yang2025thinking}. Applied to proprietary models such as GPT-4.1 and Gemini 2.5 Pro, Think3D consistently improves performance without any additional training. We further introduce Think3D-RL to teach smaller open-weight models how to use 3D tools effectively. Using only terminal answer and format rewards, without process supervision or handcrafted exploration trajectories, Think3D-RL enables Qwen3-VL-4B~\cite{bai2025qwen3} to autonomously learn effective 3D exploration strategies. After training, the model develops tool-use patterns similar to those of stronger proprietary models, while shifting the effect of 3D tool use on MindCube-1K from a (2.5\%) performance drop to a (3.58\%) improvement.

Our main contributions can be summarized as follows:

\begin{itemize}
    \item \textbf{A new perspective on spatial reasoning.}
We frame spatial reasoning as an active 3D exploration process, referred to as \emph{Think with 3D Space}, rather than conventional passive 2D perception.

    \item \textbf{A framework for explicit 3D interaction.}
    We design Think3D, allowing the VLM-based agent to manipulate point clouds through camera-based reference actions and iterative spatial reasoning chains.

    \item \textbf{Reinforcement learning for spatial exploration.}
    We formulate the model’s acquisition of viewpoint and action selection as an RL process, enabling it to develop efficient 3D exploration strategies that enhance reasoning performance across spatial benchmarks.
\end{itemize}

\section{Related Work}

\subsection{VLMs for Spatial Reasoning}

Recent VLMs improve spatial reasoning through 3D reconstruction, depth cues, spatial data, visual grounding, and explicit reasoning~\cite{fan2025vlm,cheng2024spatialrgpt,chen2024spatialvlm,roy2025bydeway,balazadeh2024synthetic,fan2025grit,wu2025reinforcing}. These capabilities have also been applied to robotics and navigation~\cite{ji2025robobrain,team2025gemini,zhou2025roborefer,zhou2024navgpt}. Some methods use code to process 3D outputs and decompose spatial tasks~\cite{chen2025geometrically,luo2026pySpatial}. MindJourney~\cite{yang2025mindjourney} explores imagined 3D space through video generation and beam search. In contrast, Think3D lets VLMs actively manipulate reconstructed point clouds, providing more reliable geometric evidence.

\subsection{VLM Tool Calling}

Tool-augmented VLMs employ external models or programs to support visual reasoning~\cite{shen2023hugginggpt,wu2023visual,suris2023vipergpt,zhao2025pyvision}. Recent methods further optimize tool-use policies through supervised learning or RL~\cite{liu2024llava,wang2025mllm,wu2025vtool,zheng2025driveagent}. For spatial reasoning, GCA~\cite{chen2026geometrically} coordinates perception and geometry tools under explicit geometric constraints, whereas SpatialClaw~\cite{cho2026spatialclaw} uses a stateful code interface to iteratively compose, inspect, and revise tool operations. pySpatial~\cite{luo2026pySpatial} generates 3D visual programs that compose calls to a predefined spatial tool API. Inspired by the ``think with images'' paradigm of DeepEyes~\cite{zheng2025deepeyes}, Think3D extends visual reasoning beyond 2D image operations by enabling active exploration of reconstructed 3D scenes.

\section{Think3D for Spatial Reasoning}
\label{sec:method}

\begin{figure*}[tbp]
    \centering
    \includegraphics[width=1.0\linewidth]{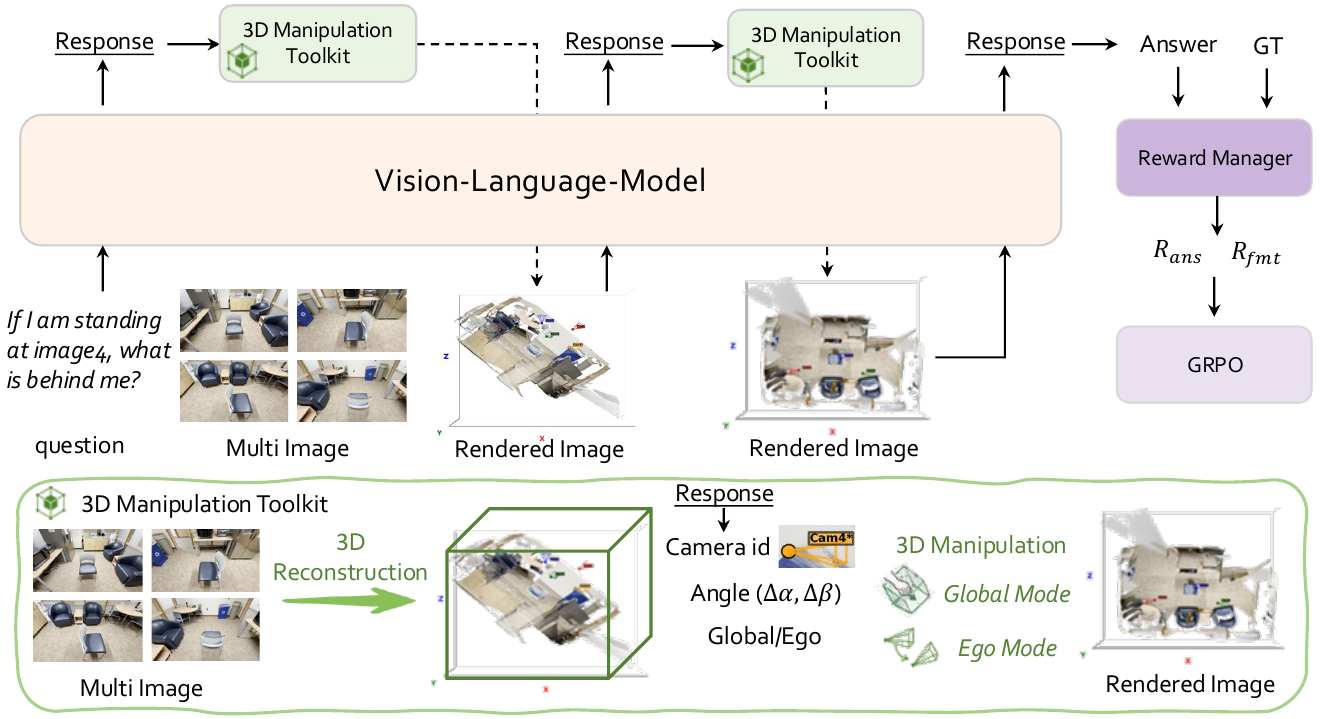}
    \caption{The \textbf{Think3D} pipeline. The VLM interacts with the 3D scene through iterative calls to the 3D Manipulation Toolkit, issuing viewpoint-manipulation actions that control camera pose and rendering parameters.
Each rendered image is appended to the agent’s memory and informs the next reasoning step, forming a repeated cycle of observe → manipulate → reflect.}
    \Description{Think3D pipeline connecting a vision-language reasoning agent to a 3D manipulation toolkit. The agent chooses camera and viewpoint parameters, receives rendered observations, and repeats the reasoning loop. Think3D-RL optimizes this exploration policy.}
    \label{fig:framework}
\end{figure*}

\subsection{Framework Overview}

As illustrated in Figure~\ref{fig:framework}, \textbf{Think3D} equips a VLM with the ability to explore and reason directly in 3D via a multi-turn \emph{observe → manipulate → reflect} loop.
Given multi-view images $\{I_t\}_{t=1}^{T}$ and a query $q$, the VLM autonomously decides whether to invoke the 3D reconstruction tool to obtain a 3D point cloud and camera poses. During the subsequent 3D interaction process, the VLM is able to iteratively manipulate the point cloud and observe the 3D environment from novel views.  
By progressively accumulating complementary geometric observations, the VLMs form an explicit \emph{3D chain of thought}, facilitating structured spatial exploration that cannot be achieved using static 2D inputs alone. 
We run the loop for at most $K$ iterations (default $K{=}3$), with at most one reconstruction call per query and at most $K$ rendering calls.
The agent can terminate early by issuing a \texttt{STOP} action when it deems the evidence sufficient.
The above 3D interaction process is powered by the following three key components of Think3D. We present the details in the subsequent sections.
\begin{itemize}
    \item \textbf{3D Manipulation Toolkit} integrates a suite of callable 3D tools, providing the agent with flexible and expressive control for exploring the 3D environment.
    \item \textbf{Spatial Reasoning Agent} performs 3D interactions by calling 3D manipulation tools and reasoning over the geometric observations.
    \item \textbf{Think3D-RL Reinforcement Learning Module} optimizes multi-step 3D exploration policy through tool calling, trained with Group Relative Policy Optimization (GRPO)~\cite{shao2024deepseekmath}.
\end{itemize}

\subsection{3D Manipulation Toolkit}
\label{sec:3d_grounding}
Under the Think3D framework, a suite of callable 3D tools enables flexible agentic 3D manipulation and exploration, featuring three core functionalities: 3D reconstruction, 3D transformation, and novel-view rendering.
\subsubsection{3D Reconstruction:} Given multi-view images $\{I_t\}_{t=1}^{T}$, a 3D point cloud and the corresponding camera poses can be estimated using Pi3~\cite{wang2025pi}.  
Each camera is represented as
\begin{equation}
    C_t = (\mathbf{K}_t, \mathbf{R}_t, \mathbf{t}_t),
\end{equation}
where $\mathbf{K}_t \in \mathbb{R}^{3\times3}$ denotes the intrinsic matrix,  
$\mathbf{R}_t \in SO(3)$ denotes the rotation matrix,  
and $\mathbf{t}_t \in \mathbb{R}^3$ represents the camera center in world coordinates. 
Here, $t$ indexes the input views.
Depth and confidence predictions are fused across views to obtain a clean colored point cloud:
\begin{equation}
    \mathcal{X} = \{(\mathbf{x}_n, \mathbf{c}_n)\}_{n=1}^{N},
\end{equation}
where $\mathbf{x}_n$ is the 3D location and $\mathbf{c}_n$ is the RGB color.

\subsubsection{3D Transformation} To enable flexible 3D exploration, the agent manipulates the reconstructed 3D point cloud to select informative viewpoints.
At each step, it predicts:
(i) a discrete camera index $i \in \{1,\dots,T\}$,
(ii) a pair of rotation angles $(\Delta\alpha, \Delta\beta)$ specifying horizontal (azimuth) and vertical (elevation) rotations,
and (iii) a binary transformation mode $m \in \{\mathrm{global}, \mathrm{ego}\}$ indicating whether to use a global or an ego-centric view.
For $m=\mathrm{global}$, we define a global center $\mathbf{c}$ as the centroid of $\mathcal{X}$ and apply a scene-centric similarity transform, i.e., we rotate (and optionally scale) the \emph{scene} around $\mathbf{c}$ while keeping the camera fixed:
\begin{equation}\label{eq:global_transform}
    \mathbf{x}_n' = s\,\Delta\mathbf{R}(\Delta\alpha,\Delta\beta)\,(\mathbf{x}_n - \mathbf{c}) + \mathbf{c},
\end{equation}
where $\Delta\mathbf{R}(\Delta\alpha,\Delta\beta)\in SO(3)$ is induced by the predicted angles and $s$ controls the zoom level of the global view (we set $s{=}1$ by default).
This yields an updated point cloud $\mathcal{X}^{\mathrm{g}} = \{(\mathbf{x}_n', \mathbf{c}_n)\}_{n=1}^{N}$.
We then render $\mathcal{X}^{\mathrm{g}}$ using the selected camera $C_i$ to provide a consistent global overview.
For $m=\mathrm{ego}$, we keep the point cloud fixed and apply a camera-centric rotation around the anchor camera center. Given the selected anchor camera $C_i$, we construct a virtual camera:
\begin{equation}\label{eq:camera}
    C_{\mathrm{new}} = (\mathbf{K}_i,\, \Delta\mathbf{R}(\Delta\alpha,\Delta\beta)\,\mathbf{R}_i,\, \mathbf{t}_i),
\end{equation}
where $\Delta\mathbf{R}(\Delta\alpha,\Delta\beta)\in SO(3)$ is an incremental rotation applied to the camera orientation, while the camera center remains fixed at $\mathbf{t}_i$.
When $\Delta\mathbf{R}=\mathbf{I}$, the virtual camera $C_{\mathrm{new}}$ reduces to the original camera $C_i$.

\subsubsection{Novel View Rendering}
In the global mode, we render the transformed point cloud $\mathcal{X}^{\mathrm{g}}$ with the anchor camera $C_i$ to obtain a global view of the 3D scene.
In the ego-centric mode, we emulate a first-person perspective by restricting $\mathcal{X}$ to a wide field-of-view cone aligned with the forward direction of the virtual camera $C_{\mathrm{new}}$, yielding $\mathcal{X}^{\mathrm{e}}$.
A lightweight, point-based renderer then produces the synthesized view:
\begin{equation}\label{eq:render}
    \hat{I}_k = \mathrm{Render}\big(\mathcal{X}^{(m)}, C^{(m)}, m\big),
\end{equation}
where $\mathcal{X}^{(m)}=\mathcal{X}^{\mathrm{g}}$ and $C^{(m)}=C_i$ for the global mode, and $\mathcal{X}^{(m)}=\mathcal{X}^{\mathrm{e}}$ and $C^{(m)}=C_{\mathrm{new}}$ for the ego mode.

\subsection{VLM-based Spatial Reasoning Agent}
As shown in Figure~\ref{fig:framework}, the VLM-based agent iteratively interacts with the 3D scene via the manipulation toolkit and accumulates rendered observations for spatial reasoning.

In the $k$-th iteration, given the history $\mathcal{H}_{k-1}$, the VLM acts as a multimodal policy:
\begin{equation}
    \mathbf{z}_k = \pi_\theta\big(q, \{I_t\}_{t=1}^{T}, \mathcal{H}_{k-1}\big),
\end{equation}
where $q$ and $\{I_t\}_{t=1}^{T}$ denote the input query and the original multi-view images, respectively.
The output $\mathbf{z}_k$ is parsed into a textual response and an optional tool call.

We optionally invoke 3D reconstruction once at the beginning with a binary decision $r\in\{0,1\}$.
When $r=1$, Pi3~\cite{wang2025pi} reconstructs the point cloud $\mathcal{X}$ and estimates camera poses $\{C_t\}_{t=1}^{T}$ from the multi-view inputs.
For viewpoint manipulation and rendering, the tool-call parameters at iteration $k$ are:
\begin{equation}
    \mathbf{a}_k = (i_k, m_k, \Delta\alpha_k, \Delta\beta_k),
\end{equation}
where $i_k \in \{1,\dots,T\}$ selects the anchor camera $C_{i_k}$.
The mode $m_k \in \{\mathrm{global}, \mathrm{ego}\}$ specifies the view type, and $\Delta\alpha_k, \Delta\beta_k$ denote the azimuth and elevation angles.

Given $\mathbf{a}_k$, the toolkit applies the transformation associated with $m_k$.
If $m_k=\mathrm{ego}$, it constructs a virtual camera via \cref{eq:camera}:
\begin{equation}
    C_{\mathrm{new}}^{(k)} = \big(\mathbf{K}_{i_k},\, \Delta\mathbf{R}(\Delta\alpha_k,\Delta\beta_k)\,\mathbf{R}_{i_k},\, \mathbf{t}_{i_k}\big).
\end{equation}
If $m_k=\mathrm{global}$, it applies the global transform in \cref{eq:global_transform} to obtain $\mathcal{X}^{\mathrm{g}}$ while keeping the anchor camera $C_{i_k}$ fixed.
The renderer then synthesizes the novel view $\hat{I}_k$ according to \cref{eq:render}, which is appended to the history:
\begin{equation}
    \mathcal{H}_k = \mathcal{H}_{k-1}\ \Vert\ \big[(\hat{I}_k, \mathbf{a}_k)\big].
\end{equation}
Thus, Think3D implements an iterative \emph{observe $\rightarrow$ manipulate $\rightarrow$ reflect} loop, where the VLM maintains an explicit 3D-aware reasoning trace over the rendered views.
The detailed prompts are provided in Appendix~\ref{app:prompts}.

\subsection{Think3D-RL for Multi-Step Exploration}

While the reasoning loop allows the model to explore 3D space, its effectiveness depends on learning \emph{which viewpoints} provide informative observations and \emph{when} such exploration should be conducted.  
We therefore optimize the exploration policy using reinforcement learning.

\subsubsection{Trajectory Formulation \& Training-time Sampling.}
We represent an agentic reasoning episode as the following trajectory:
\begin{equation}
    \tau = \{(\mathbf{s}_1, \mathbf{o}_1), (\mathbf{s}_2, \mathbf{o}_2)\dots,(\mathbf{s}_K, \mathbf{o}_K),\hat{y}\},
\end{equation}
where $\mathbf{s}_k=(q, \{{\mathbf{I}_t}\}, \mathcal{H}_{k-1})$ represents an input to the VLM agent at the $k$-th iteration; $\hat{y}$ denotes the final answer generated by the agent; and $K$ denotes the total number of exploration steps determined by the agent.

To avoid repeated 3D reconstruction during optimization, we precompute a point cloud for each sample offline. At the $k$-th exploration step, the agent first decides whether to invoke the tool. If the tool is invoked, the agent specifies the anchor camera, view mode, and rotation angles. We then render the requested view from the precomputed point cloud as the observation $\mathbf{o}_k$. This design preserves a fully interactive reasoning loop while enabling efficient training and evaluation.

\subsubsection{Outcome-Level Reward}

We assign supervision only after the model produces its terminal answer:
\begin{equation}
    R(\tau)
    =
    R_{\mathrm{ans}}(\hat{y})
    +
    \lambda_{\mathrm{fmt}}R_{\mathrm{fmt}}(\hat{y}),
\end{equation}
where $R_{\mathrm{ans}}$ equals one if the predicted option matches the
ground-truth answer and zero otherwise.
$R_{\mathrm{fmt}}$ encourages a valid answer format, and
$\lambda_{\mathrm{fmt}} = 1$.
No intermediate reward is provided for camera selection, viewing angle,
interaction length, or rendered observations.
Consequently, the policy must discover useful exploration behavior from
the terminal task outcome.

\subsubsection{Optimization.}
We train the policy with Group Relative Policy Optimization (GRPO)~\cite{shao2024deepseekmath} for stable multi-turn reasoning. Following \cite{zheng2025deepeyes}, we use a token-wise mask to stop gradients on observation tokens, optimizing only model-generated action and answer tokens.

\subsection{Prompting and Execution Workflow}
\label{sec:execution}
The interaction is organized through a system prompt, a tool prompt, and
a continuation prompt. The system prompt defines the reasoning workflow
and the required output format. The tool prompt describes the reconstruction
and rendering interface, including camera selection, rotation angles, and
the ego-centric view option. The continuation prompt supplies the current
question, previous response, tool-execution summary, original images, and
available rendered views at each reasoning step. This modular organization
keeps the task and the accumulated spatial evidence available during
multi-round exploration.

\paragraph{Viewpoint specification.}
The input camera numbering follows the image order, so selecting camera
$i$ anchors the manipulation to the $i$-th input image. The prompts explain
that the original input already provides the default $(0^\circ,0^\circ)$
view and encourage requesting complementary observations. They describe
left, right, top, back, and diagonal views, together with the option to
inspect the scene from a selected camera's position. The agent uses these
tool arguments to request views relevant to the question and incorporates
the returned images into subsequent reasoning.

\paragraph{Training workflow.}
For RL, we reconstruct the point cloud for each training scene offline
using Pi3 and reuse it throughout policy optimization. At each exploration
step, the agent predicts viewpoint parameters, including the anchor camera,
view mode, and rotation angles, conditioned on the question and the
accumulated observations. The toolkit applies these parameters to the
precomputed point cloud and renders the requested image on demand. This
image is returned to the agent as the next observation, keeping viewpoint
selection and spatial reasoning interactive while avoiding repeated
reconstruction during rollouts.
The continuation prompt distinguishes intermediate exploration from the
final answer stage. Intermediate turns may request another view or provide
an answer, whereas the final-turn prompt requests an answer using the
original inputs and the rendered observations already obtained. The
complete prompts and training-time viewpoint settings are provided in Appendix~\ref{app:prompts}.

\section{Experiment}

\subsection{Experiment Setup}
\subsubsection{Setting and Dataset}
Our Think3D-RL is implemented based on SWIFT~\cite{zhao2024swiftascalablelightweightinfrastructure}. We fine-tune the VLM using the GRPO with 8 rollouts per step to estimate advantages. The model is trained for one epoch on 8 H200 GPUs with a batch size of 8 and gradient accumulation of 4, using a cosine learning rate schedule with 5\% warmup and a base learning rate of $1\times10^{-6}$. The maximum completion length is set to 1024 tokens. During training, the language model is fully fine-tuned while the vision encoder is frozen. The training set contains 977 samples randomly selected from the MindCube training dataset, with no overlap with the test set. For the standard GRPO, we adapt the same
setting. During inference, we deploy a Pi3 tool on a RTX 3090 GPU to perform inference. We provide full implementation details of Think3D and Think3D-RL in Appendix~\ref{app:prompts}.

Each setting is evaluated three times, and we report the average performance.
We use AdamW with weight decay $0.1$, a GRPO clipping parameter of $0.2$,
and a KL penalty weight of $0.05$. Training uses temperature $0.6$, and
evaluation uses temperature $1.0$. The vision encoder remains frozen during
optimization; gradient checkpointing, mixed precision, and DeepSpeed are
enabled. Further training and evaluation parameters are listed in
Appendix~\ref{app:training}.

\subsubsection{Benchmarks}
We evaluate our method on 3 challenging spatial reasoning benchmarks: BLINK~(Multi-view)~\cite{fu2024blink}, MindCube 1K~\cite{yin2025spatial}, and the video-based VSI-Bench Tiny~\cite{yang2025thinking}.
BLINK~(Multi-view) uses all the multi-view data from the BLINK dataset and focuses on multi-view geometric understanding, particularly assessing a model’s ability to infer relative camera motion across views.
MindCube 1K contains 3 canonical camera-motion types—rotation, around, and among. 
VSI-Bench Tiny assesses visual–spatial intelligence in dynamic egocentric videos across four tasks: route planning, object relative direction prediction, appearance order reasoning, and relative distance. We adopt sample 7 frames from each video for evaluation. All models are evaluated on the same sample sets for fair comparison.

\subsubsection{Baseline Models}
We compare Think3D with specialized spatial models, including SpatialLadder-3B~\cite{li2025spatialladder}, Spatial-MLLM-4B~\cite{wu2026spatial}, and VST-7B-SFT~\cite{yang2025visual}, as well as spatial tool-augmented methods, including Tiger~\cite{han2025tiger}, PySpatial~\cite{luo2026pySpatial}, and GCA~\cite{chen2026geometrically}. The general-purpose baseline models include GPT-4.1, Gemini-2.5-Pro, and Qwen3-VL-4B-Instruct. Qwen3-VL-4B\textsubscript{T3RL} denotes the model trained using Think3D-RL. Qwen3-VL-4B$_{\mathrm{GRPO}}$ refers to finetuned with standard GRPO setting.

\subsection{Main Results}
Results on the MindCube 1K~(Table~\ref{tab:mindcube}) show that Think3D substantially improves proprietary models such as GPT-4.1 and Gemini-2.5-Pro, yielding absolute gains of 7.67 and 20.20 percentage points, respectively.
In contrast, for smaller models such as Qwen3-VL-4B-Instruct, Think3D decreases the result~(-2.5\%), suggesting limited spatial reasoning capacity constrains the benefit of exploration.
However, once Qwen3-VL-4B-Instruct is fine-tuned with Think3D-RL~(Qwen3-VL-4B\textsubscript{T3RL}), it improves by 3.58\% with Think3D. By comparison, standard GRPO improves MindCube accuracy by only 0.02 percentage points.
This provides evidence that RL strengthens viewpoint selection and spatial exploration.
We further analyze how RL-trained models achieve these gains in Ablation.

\begin{table}[!htbp]
    \centering
    \small
    \caption{
    MindCube 1K accuracy (\%). Overall is recomputed from the reported
    Among/Rotation/Around scores with weights $3:1:2$.
    Parentheses show absolute changes; bold marks the better result
    within each backbone pair.
    }
    \label{tab:mindcube}

    \setlength{\tabcolsep}{6pt}
    \renewcommand{\arraystretch}{1.08}

    \fitwidth{%
    \begin{tabular*}{\linewidth}{@{\extracolsep{\fill}}lcccc@{}}
        \toprule
        \textbf{Method}
        & \textbf{Among}
        & \textbf{Rotation}
        & \textbf{Around}
        & \textbf{Overall} \\
        \midrule

        \multicolumn{5}{@{}l}{\textbf{Specialized spatial models}} \\

        SpatialLadder-3B
        & 43.2 & 35.0 & 50.8 & 44.37 \\

        Spatial-MLLM-4B
        & 30.5 & 39.0 & 36.0 & 33.75 \\

        VST-7B-SFT
        & 35.9 & 37.0 & 50.8 & 41.05 \\

        \midrule
        \addlinespace[2pt]
        \multicolumn{5}{@{}l}{\textbf{Tool-augmented models}} \\

        Tiger
        & 26.7 & 33.0 & 28.3 & 28.28 \\

        PySpatial
        & 64.9 & 41.8 & 72.7 & 63.65 \\

        GCA
        & 59.8 & 82.0 & 61.8 & 64.17 \\

        \midrule
        \multicolumn{5}{@{}l}{\textbf{General-purpose VLMs}} \\

        \mbox{Gemini 2.5 Pro}
        & 26.00
        & 85.50
        & 38.40
        & 40.05 \\

        \thinkmodel{Gemini 2.5 Pro}
        & \gaincell{49.67}{+23.67}
        & \gaincell{87.00}{+1.50}
        & \gaincell{62.75}{+24.35}
        & \gaincell{60.25}{+20.20} \\

        \midrule
        \addlinespace[3pt]

        \mbox{GPT-4.1}
        & 43.50
        & 60.00
        & 47.50
        & 47.58 \\

        \thinkmodel{GPT-4.1}
        & \gaincell{51.17}{+7.67}
        & \gaincell{63.00}{+3.00}
        & \gaincell{57.50}{+10.00}
        & \gaincell{55.25}{+7.67} \\

        \midrule
        \addlinespace[3pt]

        \mbox{Qwen3-VL-4B Instruct}
        & 27.83
        & 30.50
        & \textbf{32.50}
        & \textbf{29.83} \\

        \thinkmodel{Qwen3-VL-4B}
        & \gaincell{30.33}{+2.50}
        & \gaincell{32.00}{+1.50}
        & \dropcell{20.50}{-12.00}
        & \dropcell{27.33}{-2.50} \\

        \midrule
        \addlinespace[3pt]

        \mbox{Qwen3-VL-4B$_{\mathrm{T3RL}}$}
        & 28.00
        & 34.50
        & 30.75
        & 30.00 \\

        \thinkmodel{Qwen3-VL-4B$_{\mathrm{T3RL}}$}
        & \gaincell{32.50}{+4.50}
        & \gaincell{39.00}{+4.50}
        & \gaincell{32.50}{+1.75}
        & \gaincell{33.58}{+3.58} \\

        \midrule
        \addlinespace[3pt]

        \mbox{Qwen3-VL-4B$_{\mathrm{GRPO}}$}
        & 26.00
        & 37.50
        & 31.80
        & 29.85 \\

        \bottomrule
    \end{tabular*}%
    }
\end{table}

The results on VSI-Bench and BLINK Multi-view, reported in Table~\ref{tab:vsi_blink_results}, provide further evidence of the effectiveness of Think3D. On VSI-Bench, Think3D improves the performance of GPT-4.1 and Gemini-2.5-Pro by 2.96\% and 6.45\%, respectively, demonstrating its effectiveness for video-based spatial reasoning. BLINK Multi-view exhibits a consistent trend. Moreover, when combined with Think3D, Qwen3-VL-4B\textsubscript{T3RL} improves from 0.8\% to 12.05\%, indicating that RL enables the model to perform more effective exploration in 3D space.  Figure~\ref{fig:angle_heatmaps} provides a qualitative example. Notably, the RL training data are sampled exclusively from MindCube, while the resulting improvements generalize to distinct benchmarks.

\begin{table}[!htbp]
    \centering
    \small
    \caption{
    Results on VSI-Bench-tiny and BLINK Multi-view (\%).
    Parentheses report absolute changes over the corresponding backbone.
    Think3D uses up to two exploration iterations for proprietary models
    and up to three for Qwen3-VL-4B.
    Bold denotes the better result within each backbone pair.
    }
    \label{tab:vsi_blink_results}

    \setlength{\tabcolsep}{4pt}
    \renewcommand{\arraystretch}{1.05}

    \fitwidth{%
    \begin{tabular*}{\linewidth}{
        @{\extracolsep{\fill}}lccccc
        !{\hspace{1.5pt}\vrule width 0.4pt\hspace{1.5pt}}
        c@{}
    }
        \toprule
        \textbf{Model}
        & \makecell{\textbf{Route}\\\textbf{Plan}}
        & \makecell{\textbf{Rel.}\\\textbf{Dir.}}
        & \makecell{\textbf{Rel.}\\\textbf{Dist.}}
        & \makecell{\textbf{App.}\\\textbf{Order}}
        & \textbf{Avg.}
        & \makecell{\textbf{BLINK}\\\textbf{MV}} \\
        \midrule

        \mbox{GPT-4.1~\cite{GPT-4.1}}
        & 40.80
        & 40.63
        & 43.30
        & 68.00
        & 48.18
        & 36.84 \\

        \thinkmodel{GPT-4.1}
        & \gaincell{45.26}{+4.46}
        & \gaincell{45.30}{+4.67}
        & \gaincell{46.00}{+2.70}
        & \samecell{68.00}
        & \gaincell{51.14}{+2.96}
        & \gaincell{63.91}{+27.07} \\

        \midrule
        \addlinespace[3pt]

        \mbox{Gemini-2.5-Pro~\cite{comanici2025gemini}}
        & 45.58
        & 28.67
        & 50.67
        & 55.73
        & 45.16
        & 47.37 \\

        \thinkmodel{Gemini-2.5-Pro}
        & \gaincell{46.93}{+1.35}
        & \gaincell{37.30}{+8.63}
        & \gaincell{54.00}{+3.33}
        & \gaincell{68.24}{+12.51}
        & \gaincell{51.61}{+6.45}
        & \gaincell{50.38}{+3.01} \\

        \midrule
        \addlinespace[3pt]

        \mbox{Qwen3-VL-4B~\cite{Qwen3-VL}}
        & \textbf{34.69}
        & 40.67
        & \textbf{35.33}
        & 42.44
        & 38.28
        & 47.87 \\

        \thinkmodel{Qwen3-VL-4B}
        & \dropcell{30.61}{-4.08}
        & \gaincell{44.00}{+3.33}
        & \dropcell{29.33}{-6.00}
        & \gaincell{52.38}{+9.94}
        & \gaincell{39.08}{+0.80}
        & \gaincell{48.62}{+0.75} \\

        \midrule
        \addlinespace[3pt]

        \mbox{Qwen3-VL-4B$_{\mathrm{T3RL}}$}
        & 27.89
        & 30.67
        & 32.00
        & 42.86
        & 33.36
        & 46.11 \\

        \thinkmodel{Qwen3-VL-4B$_{\mathrm{T3RL}}$}
        & \gaincell{36.73}{+8.84}
        & \gaincell{39.00}{+8.33}
        & \gaincell{44.67}{+12.67}
        & \gaincell{61.22}{+18.36}
        & \gaincell{45.41}{+12.05}
        & \gaincell{53.39}{+7.28} \\

        \midrule
        \addlinespace[3pt]

        \mbox{Qwen3-VL-4B$_{\mathrm{GRPO}}$}
        & 28.57
        & 38.00
        & 36.00
        & 30.61
        & 33.30
        & 52.38 \\

        \bottomrule
    \end{tabular*}%
    }
\end{table}

Table~\ref{tab:training_configuration} summarizes the optimization and
execution settings used for Think3D-RL. Point clouds are prepared offline,
while the rendering step returns observations for the agent's requested
viewpoints during each rollout.

\begin{table}[!htbp]
\centering
\caption{Optimization and execution settings for Think3D-RL.}
\label{tab:training_configuration}
\begin{tabular}{@{}ll@{}}
\toprule
Parameter & Setting \\
\midrule
Backbone & Qwen3-VL-4B-Instruct \\
Training examples & 977 \\
Training epochs & 1 \\
Rollouts per prompt & 8 \\
Training batch size & 8 \\
Gradient accumulation & 4 \\
Optimizer & AdamW \\
Learning rate & $1\times10^{-6}$ \\
Learning-rate schedule & Cosine, 5\% warmup \\
Weight decay & 0.1 \\
GRPO clipping & 0.2 \\
KL penalty weight & 0.05 \\
Temperature (train / eval) & 0.6 / 1.0 \\
Maximum response length & 1024 tokens \\
Training hardware & $8\times$ H200 GPUs \\
\bottomrule
\end{tabular}
\end{table}

\FloatBarrier
\section{Ablation Study}

\subsection{Ablation of Components}

As shown in Table~\ref{tab:3d_ablation}, reconstruction alone improves BLINK
accuracy from $36.84$ to $41.17$ and MindCube accuracy from $47.58$ to $50.16$.
Adding camera anchors, camera selection, and ego-centric views further
increases the scores to $63.91$ and $55.25$, respectively. These results
highlight the contribution of controllable geometric observations to
spatial reasoning.

On BLINK, introducing camera anchors raises accuracy from $41.17$ to
$55.46$. Enabling camera selection then increases accuracy to $61.65$,
and adding ego-centric views brings it to $63.91$. On MindCube, the
corresponding scores are $50.16$, $51.90$, $52.46$, and $55.25$.
The camera anchor defines the reference frame for a manipulation, camera
selection lets the agent choose which input viewpoint to use as that
reference, and the ego-centric mode provides an observation from the
selected camera's position. The successive configurations therefore
examine complementary controls over the geometric evidence returned
to the reasoning agent.

\begin{table}[!htbp]
\centering
\caption{Ablation on different 3D reasoning components. All results are reported as accuracy (\%). 3D Rec. denotes reasoning with reconstructed 3D geometry; Cam. Anchor uses the camera pose as the manipulation anchor; Cam. Cho. enables camera selection; and Ego-view indicates whether the model may request ego-centric views. We report results on BLINK (multi-view) and MindCube-1K.}
\label{tab:3d_ablation}
\begin{tabular}{cccccc}
\toprule
\makecell{\textbf{3D}\\\textbf{Rec.}} &
\makecell{\textbf{Cam.}\\\textbf{Anchor}} &
\makecell{\textbf{Cam.}\\\textbf{Cho.}} &
\makecell{\textbf{Ego}\\\textbf{View}} &
\textbf{BLINK} & \textbf{MindCube} \\
\midrule
 &  &  &  & 36.84 & 47.58 \\
\cmark &  &  &  & 41.17 & 50.16 \\
\cmark & \cmark &  &  & 55.46 & 51.90 \\
\cmark & \cmark & \cmark &  & 61.65 & 52.46 \\
\cmark & \cmark & \cmark & \cmark & \textbf{63.91} & \textbf{55.25} \\
\bottomrule
\end{tabular}
\end{table}

\subsection{Ablation of Space Exploration Strategy}
\label{sec:space exploration strategy}

\begin{figure}[tbp]
    \centering
    \begin{subfigure}[t]{0.5\linewidth}
        \centering
        \includegraphics[width=\linewidth]{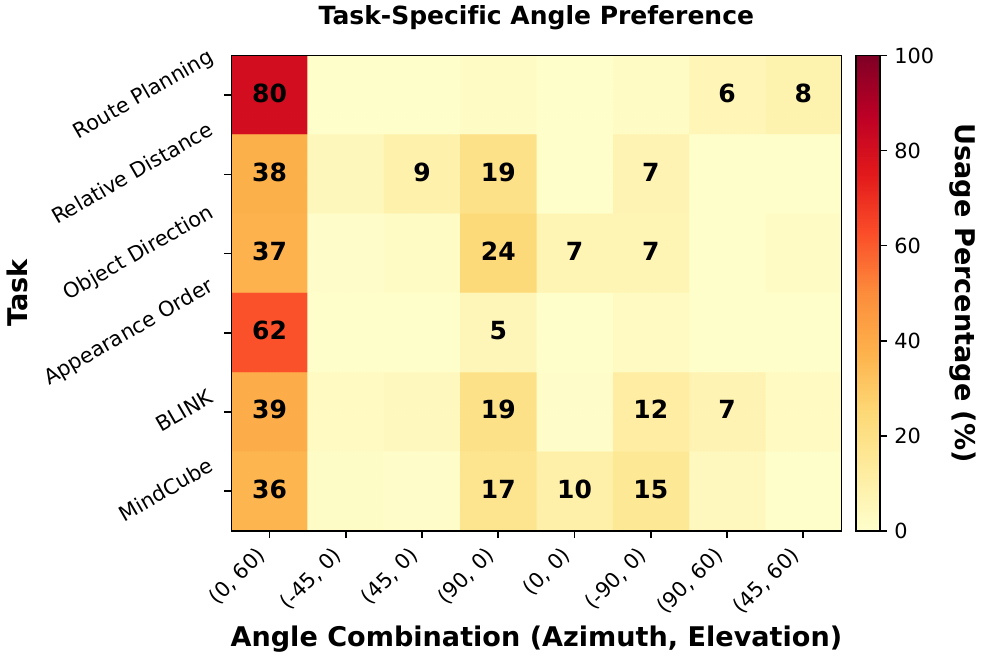}
        \caption{Task-level patterns.}
        \label{fig:task_compare}
    \end{subfigure}\hfill
    \begin{subfigure}[t]{0.5\linewidth}
        \centering
        \includegraphics[width=\linewidth]{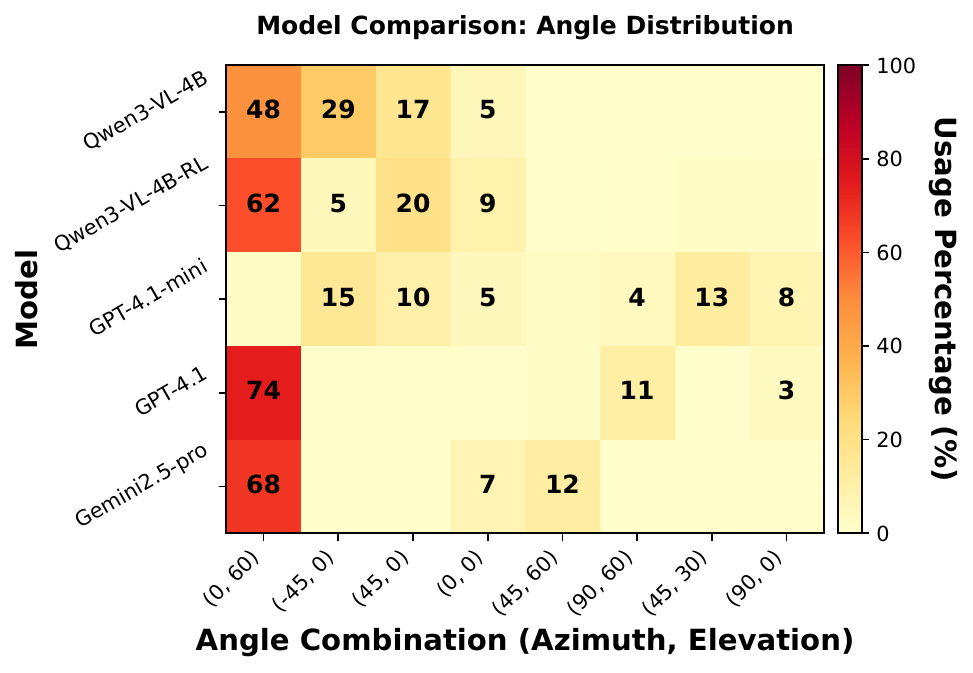}
        \caption{Model-level patterns.}
        \label{fig:model_compare}
    \end{subfigure}
    \caption{Spatial exploration patterns in viewpoint selection. Strong models concentrate on informative viewing angles, such as oblique and top-down views. After RL fine-tuning, the viewpoint distribution of Qwen3-VL-4B\textsubscript{T3RL} shifts toward a similar pattern. Exploration strategies also vary substantially across tasks; for example, route planning favors top-down views near $(0,60)$. The distributions summarize the viewpoints selected through the tool interface.}
    \Description{Two heatmaps summarize viewing-angle selection across spatial tasks and across models. The patterns compare task-dependent exploration and the change in the small model after reinforcement learning.}
    \label{fig:angle_heatmaps}
\end{figure}

As shown in Figure~\ref{fig:task_compare}, we analyze the spatial exploration strategies adopted by VLMs across multiple task categories, including multi-view reasoning, route planning, and object-orientation estimation, as well as across models with different underlying capabilities. GPT-4.1 exhibits clear task-dependent exploration patterns. For route-planning and appearance-ordering tasks, it predominantly selects top-down viewpoints to capture the global spatial layout and relative arrangement of objects. In contrast, for tasks such as MindCube and object-orientation estimation, the model more frequently explores rotational viewpoints that reveal complementary geometric cues for orientation inference. We compare random, fixed, and learned viewpoint-selection strategies below.

\subsection{Ablation of Reinforcement Learning Dynamics}
\begin{figure}[tbp]
    \centering
    \includegraphics[width=0.82\linewidth]{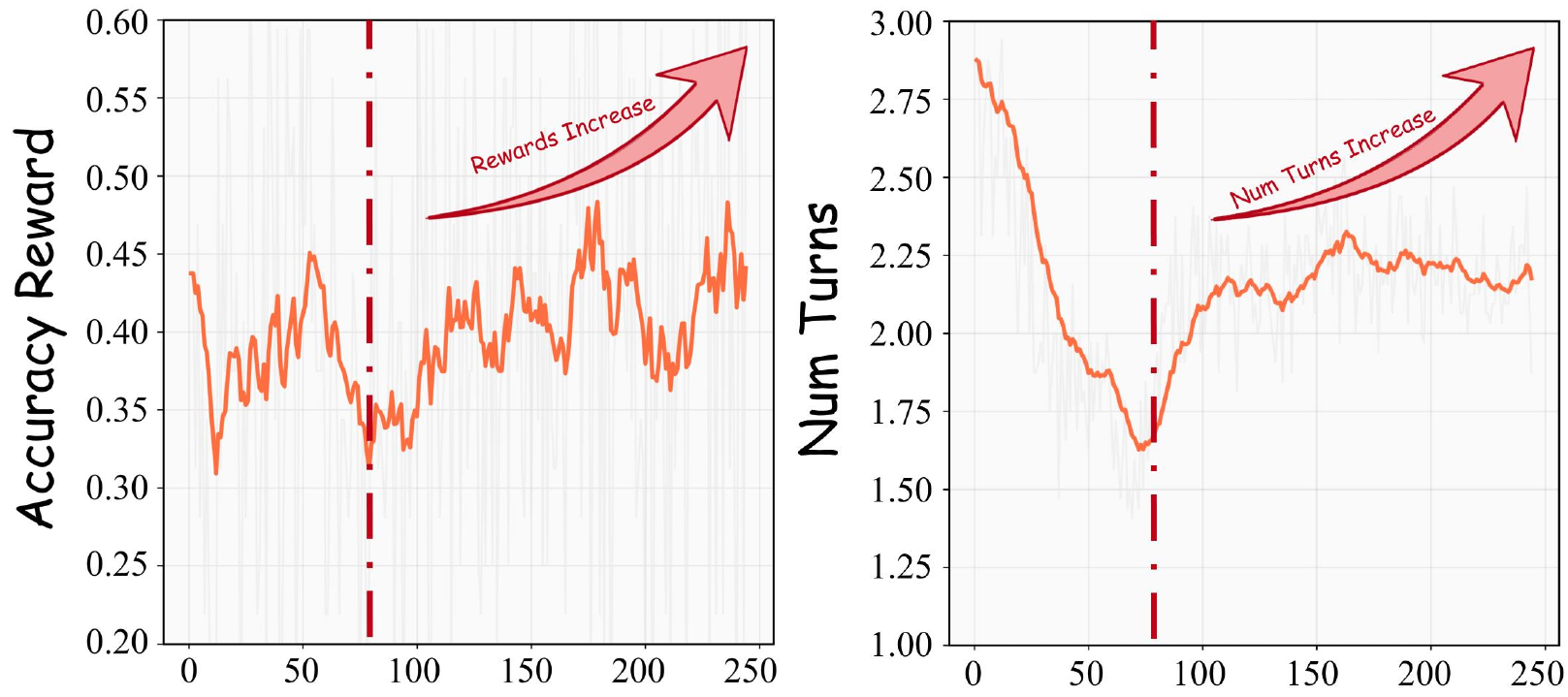}
    \caption{Reinforcement Learning Dynamics. As RL fine-tuning progresses, the model learns when extra 3D tool calls are worthwhile, shifting from shorter but less accurate trajectories to more informative explorations with higher reward.}
    \Description{Training curves show the evolution of accuracy reward and the number of reasoning turns during reinforcement learning.}
    \label{fig:rl_dynamic}
\end{figure}

As shown in Figure~\ref{fig:rl_dynamic}, we visualize RL training dynamics of one checkpoint by tracking the evolution of the accuracy reward and the number of turns per trajectory.
During the first 50 training steps, the model tends to reduce turns to increase reward. However, this reduction causes a noticeable drop in accuracy: with fewer turns, the model invokes spatial tools less often and thus obtains fewer 3D viewpoints.
After about 50 training steps, the model gradually increases its spatial tool usage to render point-cloud images, resulting in steady improvement in accuracy.

\subsection{Ablation on What the Model Learns through RL}
\label{ab:whatrllearn}

To better understand what the model learns from reinforcement learning, we analyze spatial exploration behavior before and after RL fine-tuning.
We visualize viewpoint distributions of strong models such as GPT-4.1 and Gemini-2.5-Pro, whose robust strategies correlate with substantial gains under Think3D.
We then compare these behaviors with those of a smaller model, Qwen3-VL-4B, and its RL-enhanced variant, Qwen3-VL-4B\textsubscript{T3RL}.
As shown in Figure~\ref{fig:model_compare}, Qwen3-VL-4B\textsubscript{T3RL} adopts viewpoint patterns closer to the stronger models, for example selecting top-down perspectives more often to capture global spatial structure.
These results indicate that RL improves informed 3D exploration.

\subsection{Ablation of Viewpoint Selection}
\label{sec:policy_controls}
We compare four exploration strategies on BLINK Multi-view. Random selects two
views from eight candidate angles, while Heuristic uses the fixed views
$(-45,30)$ and $(45,30)$. Both use the original Qwen3-VL-4B backbone.
RL$_{\mathrm{random}}$ uses the RL-trained backbone with random exploration,
and RL uses the learned exploration policy. All methods run three rounds.

\begin{table}[!htbp]
\centering
\caption{Effect of viewpoint-selection strategies on BLINK Multi-view.
All results are accuracy (\%).}
\label{tab:exp_diff_angle_choose}
\begin{tabular}{lc}
\toprule
Exploration strategy & BLINK Multi-view \\
\midrule
Random & 45.86 \\
Heuristic & 47.37 \\
RL$_{\mathrm{random}}$ & 45.11 \\
RL & \textbf{53.39} \\
\bottomrule
\end{tabular}
\end{table}

As shown in Table~\ref{tab:exp_diff_angle_choose}, random exploration with
the RL-trained backbone yields $45.11$, compared with $53.39$ under the
learned policy. Fixed heuristic viewpoints reach $47.37$. These comparisons
support the benefit of learning how to select useful spatial observations,
beyond simply supplying additional rendered views.

\subsection{Task-Dependent Tool Use}
\begin{figure}[!htbp]
\centering
\begin{subfigure}[t]{0.485\linewidth}
\centering
\includegraphics[width=\linewidth]{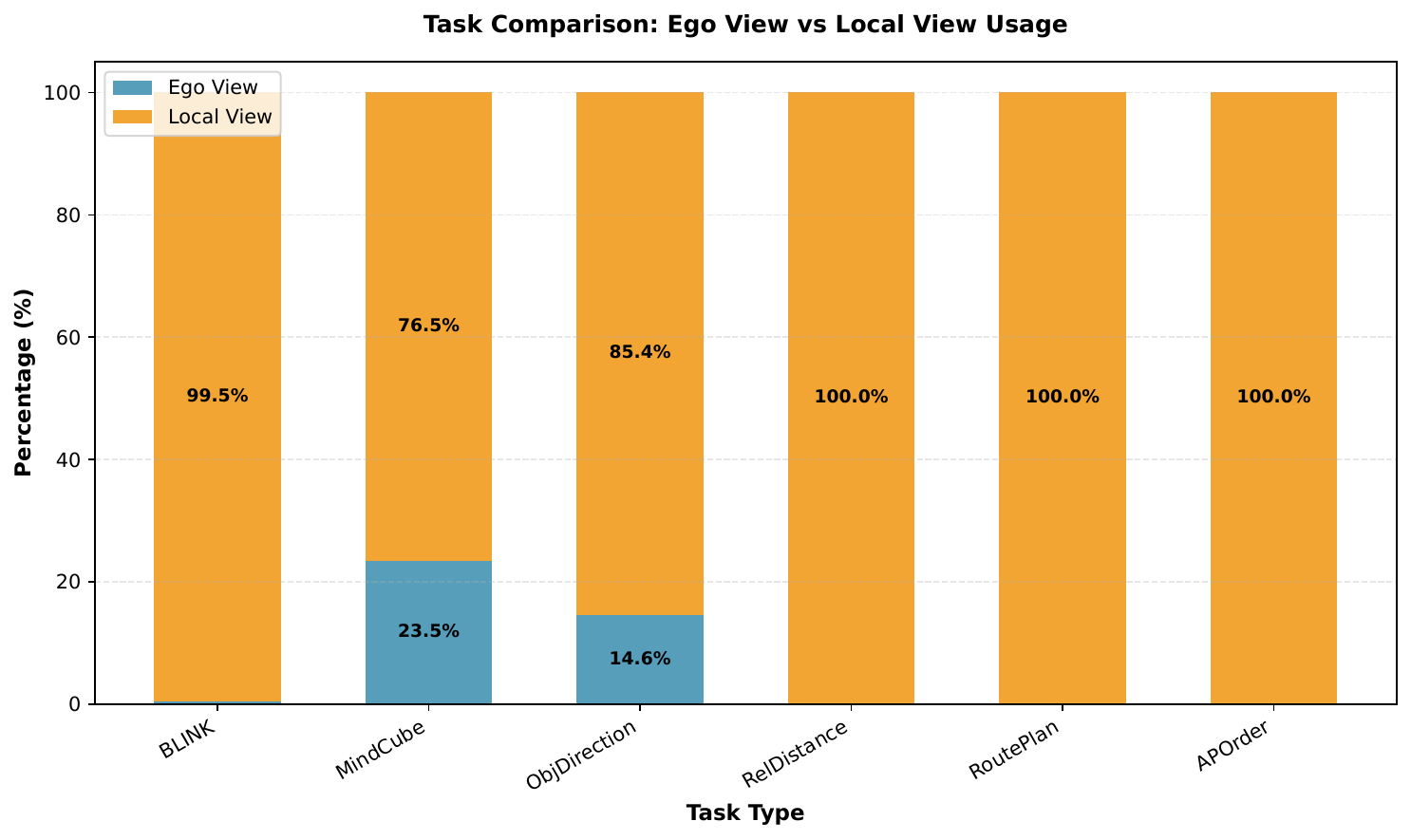}
\caption{View-mode usage.}
\end{subfigure}\hfill
\begin{subfigure}[t]{0.485\linewidth}
\centering
\includegraphics[width=\linewidth]{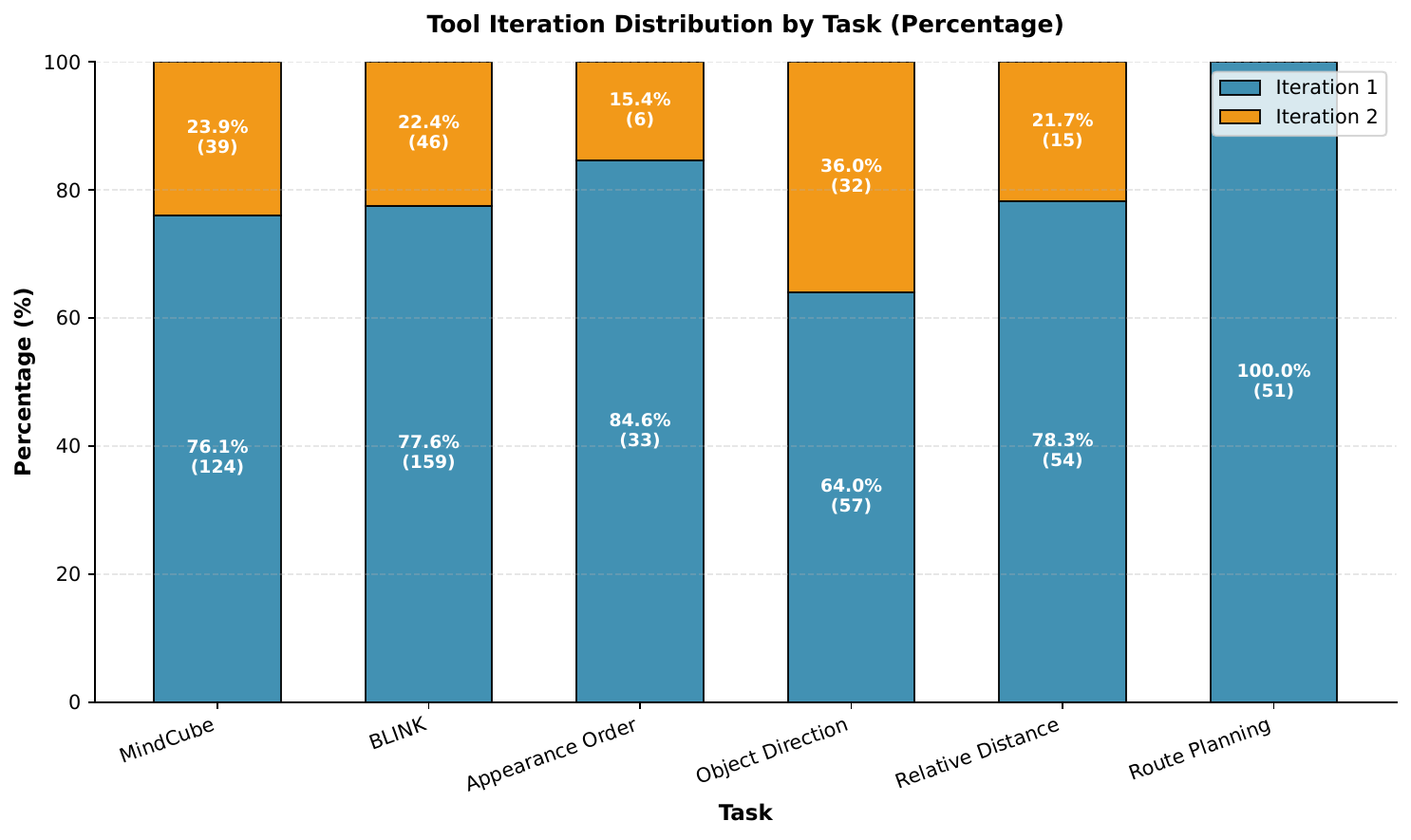}
\caption{Tool-call iterations.}
\end{subfigure}
\caption{GPT-4.1's view-mode usage and tool-call iteration distributions
across spatial reasoning tasks.}
\Description{Two bar charts show GPT-4.1's view-mode usage and the number of tool-call iterations across spatial reasoning tasks.}
\label{fig:task_policy}
\end{figure}
Figure~\ref{fig:task_policy} reports how GPT-4.1 uses the tool across different
tasks. MindCube and object-direction questions exhibit a higher proportion
of ego-centric views, which provide observations from a selected camera's
position. Route planning rarely uses ego-centric views. These choices
complement the task-dependent angle distributions in
Figure~\ref{fig:task_compare}: different questions lead the agent to inspect
different aspects of the reconstructed scene.

\begin{figure*}[t]
    \centering
    \includegraphics[width=\linewidth]{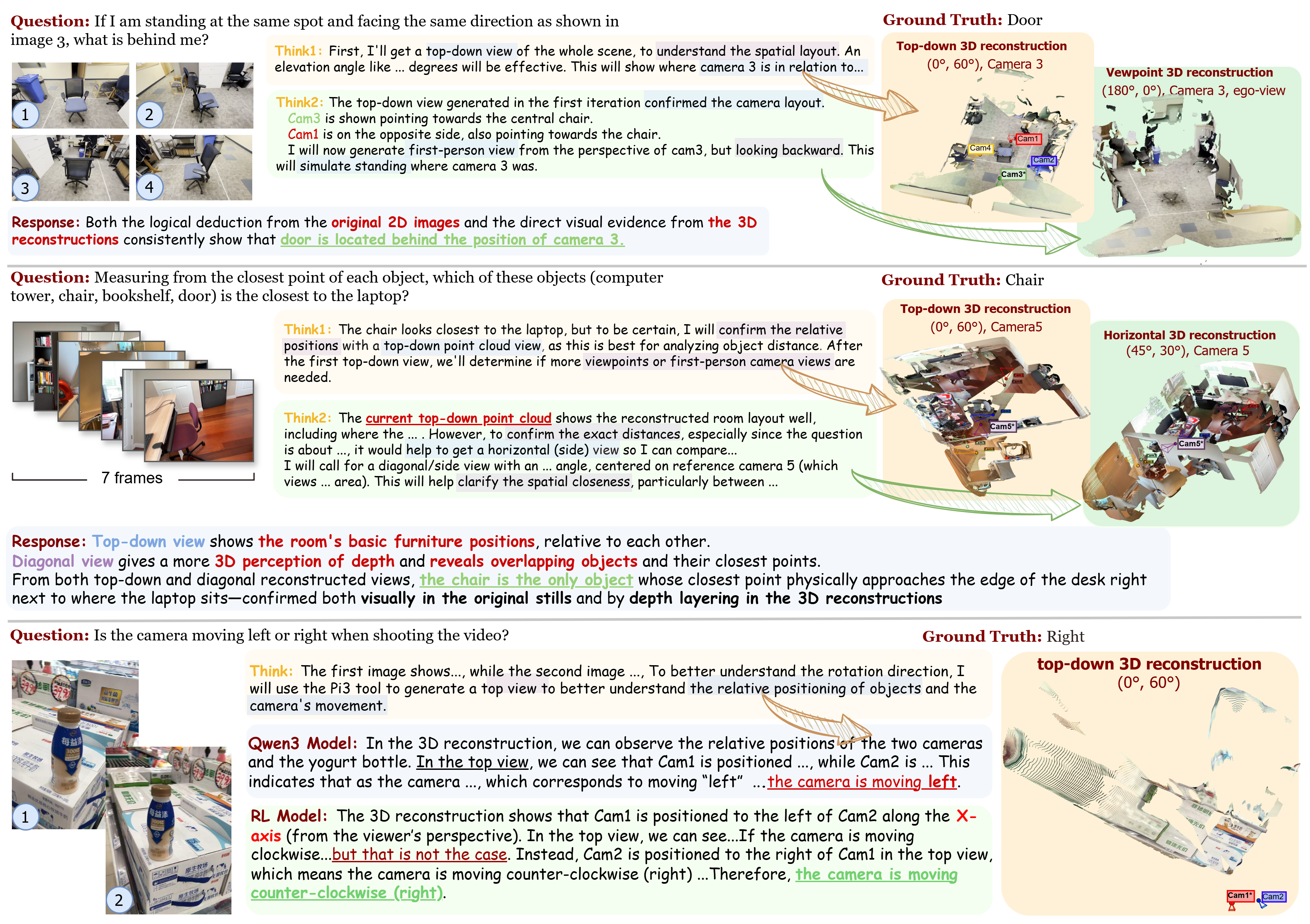}
    \caption{Spatial exploration behavior of Think3D. The agent autonomously selects viewpoints and switches between global and ego-centric views; after RL training, it explores angles more systematically than the untuned baseline.}
    \Description{Three spatial reasoning examples show input images, selected reconstructed views, intermediate reasoning, and answers. The final example compares the base model with the model trained using reinforcement learning.}
    \label{fig:example}
\end{figure*}

Tool-call iteration counts also vary across tasks. Route-planning questions
use fewer tool-call iterations, whereas the other task categories include
more multi-round exploration. The view-mode and iteration distributions
therefore provide complementary descriptions of the spatial exploration
process: the agent selects both the form of its next observation and how
many rounds of interaction to perform.

\subsection{Qualitative Spatial Exploration}
Figure~\ref{fig:example} illustrates the interaction process on three spatial
reasoning questions. In the first example, the question asks which object
lies behind the observer at the position and orientation of the third input
camera. The agent first requests a top-down reconstruction to inspect the
scene layout, then switches to an ego-centric view referenced to that
camera. Combining the global arrangement with the camera-relative
observation helps the agent identify the door behind the observer.

The second example concerns the object whose closest point is nearest to
the laptop. The agent first inspects an overhead view of the room and then
requests an oblique view to examine the relative positions and depth of
the candidate objects. It combines these geometric observations with the
original video frames to identify the chair. This sequence illustrates how
different viewpoints supply complementary evidence for a distance question.

The third example concerns whether the camera moves left or right between
two images. The reconstruction makes the relative positions of the input
cameras explicit. The untuned and RL-trained models interpret the same
spatial arrangement differently; the RL-trained model uses the reconstructed
camera positions to identify the motion to the right. Together, these
examples illustrate the use of camera anchors, viewpoint changes, and
rendered geometric observations within the multi-round reasoning process.

\section{Conclusion}
We introduce Think3D, a framework that enables VLM agents to actively
explore reconstructed 3D scenes for spatial reasoning. Camera-anchored
manipulation and iterative rendering provide complementary geometric
observations, while Think3D-RL trains smaller models to select useful
views through terminal answer and format rewards. Experiments on BLINK,
MindCube, and VSI-Bench-Tiny demonstrate improvements over the corresponding
backbones. The component ablations, viewpoint-selection comparisons, and
task-dependent tool-use analysis further characterize how the agent
interacts with 3D space. These results support explicit geometric
interaction as an effective approach to spatial reasoning in VLM agents.

\FloatBarrier
\pdfbookmark[1]{References}{references}
\bibliographystyle{plainnat}
\bibliography{references}

\clearpage
\appendix

\section{Prompts and Implementation Details}
\label{app:prompts}
\begin{figure*}[tbp]
    \centering
    \includegraphics[width=\linewidth,height=0.88\textheight,keepaspectratio]{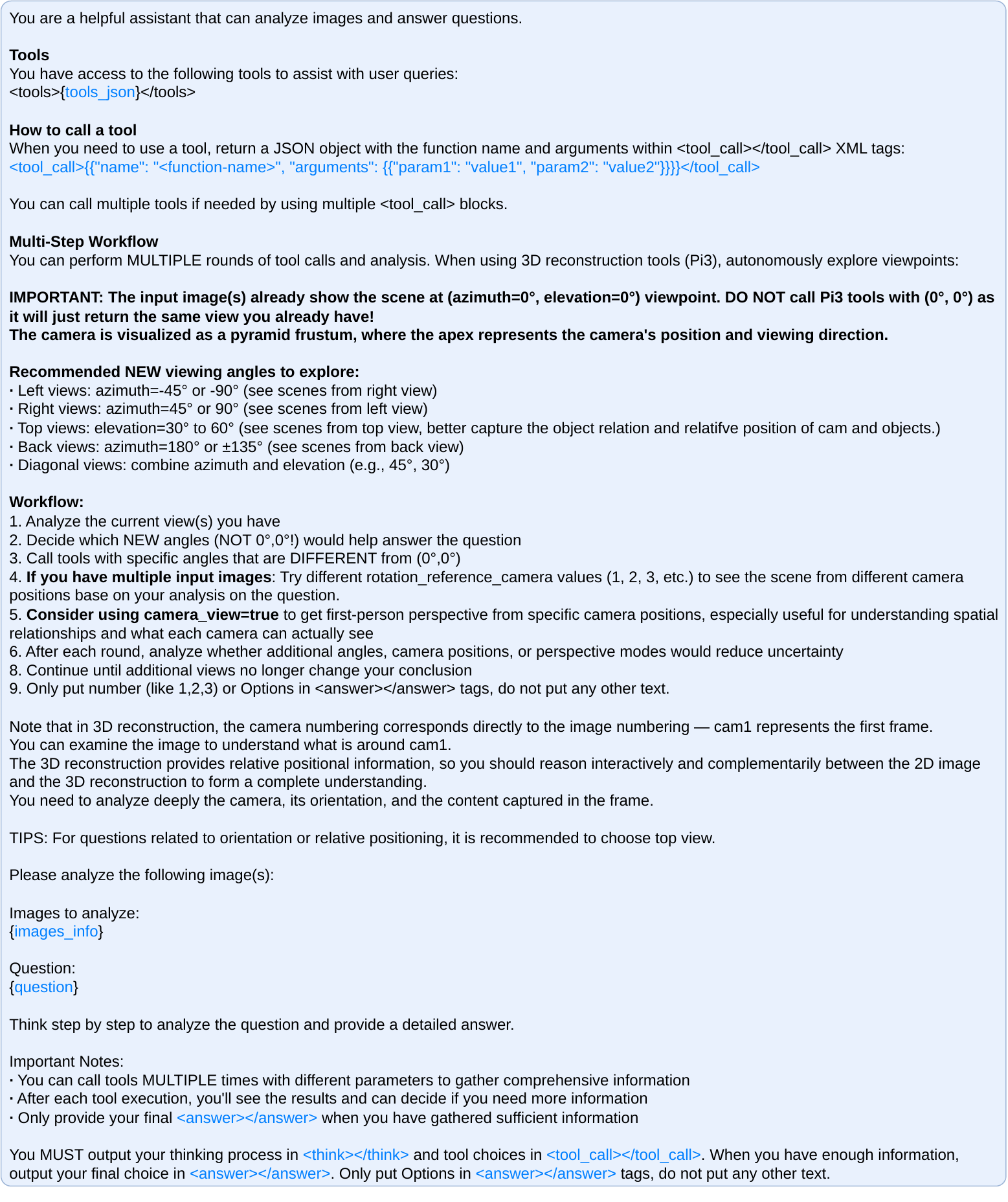}
    \caption{The system prompt. Instruction prompt detailing tool invocation rules and the multi-step workflow for iterative 3D viewpoint exploration, including tool-call format, recommended angles, and guidelines for reasoning with reconstructed camera poses.}
    \Description{Full system prompt specifying the reasoning workflow, 3D tool calls, viewpoint exploration rules, camera references, and answer format.}
    \label{fig:system_prompt}
\end{figure*}

\begin{figure}[tbp]
    \centering
    \includegraphics[width=\linewidth,height=0.88\textheight,keepaspectratio]{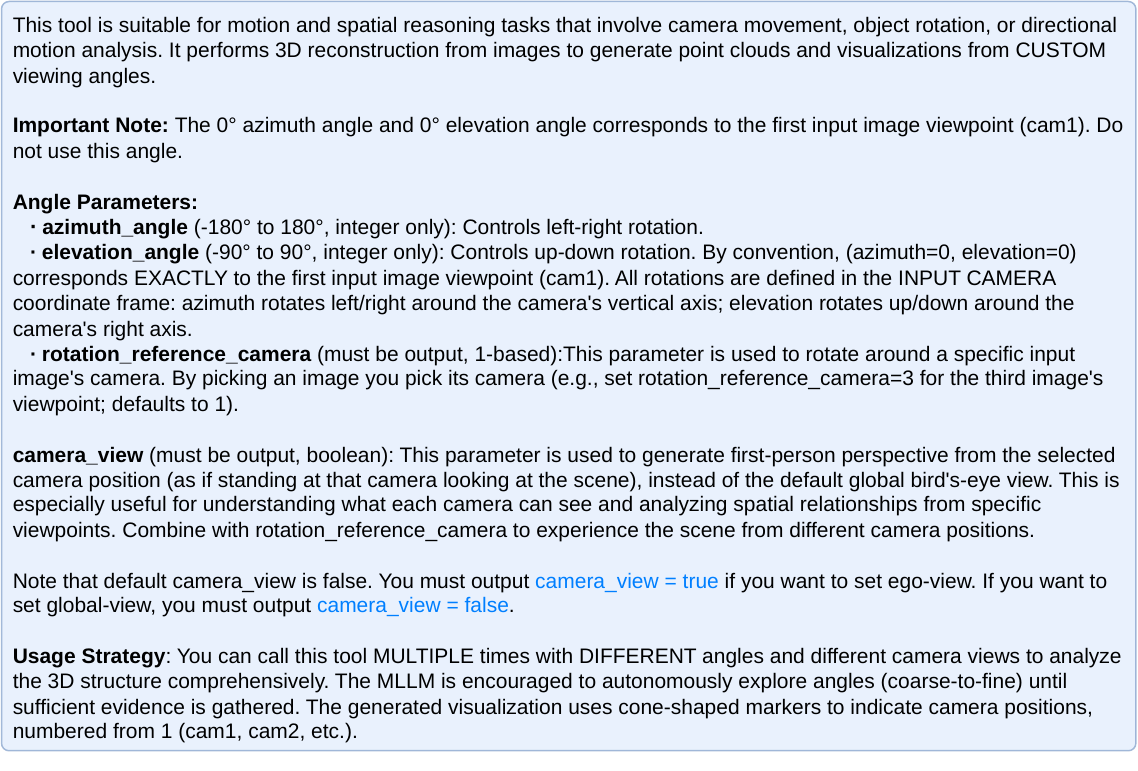}
    \caption{The Pi3 Tool Prompt. The prompt specifies the tool's capabilities, key control parameters, and multi-angle query usage strategies to support comprehensive spatial understanding.}
    \Description{Pi3 tool prompt listing tool capabilities, control parameters, and instructions for requesting multiple viewpoints.}
    \label{fig:pi3_prompt}
\end{figure}

\begin{figure}[tbp]
    \centering
    \includegraphics[width=\linewidth,height=0.88\textheight,keepaspectratio]{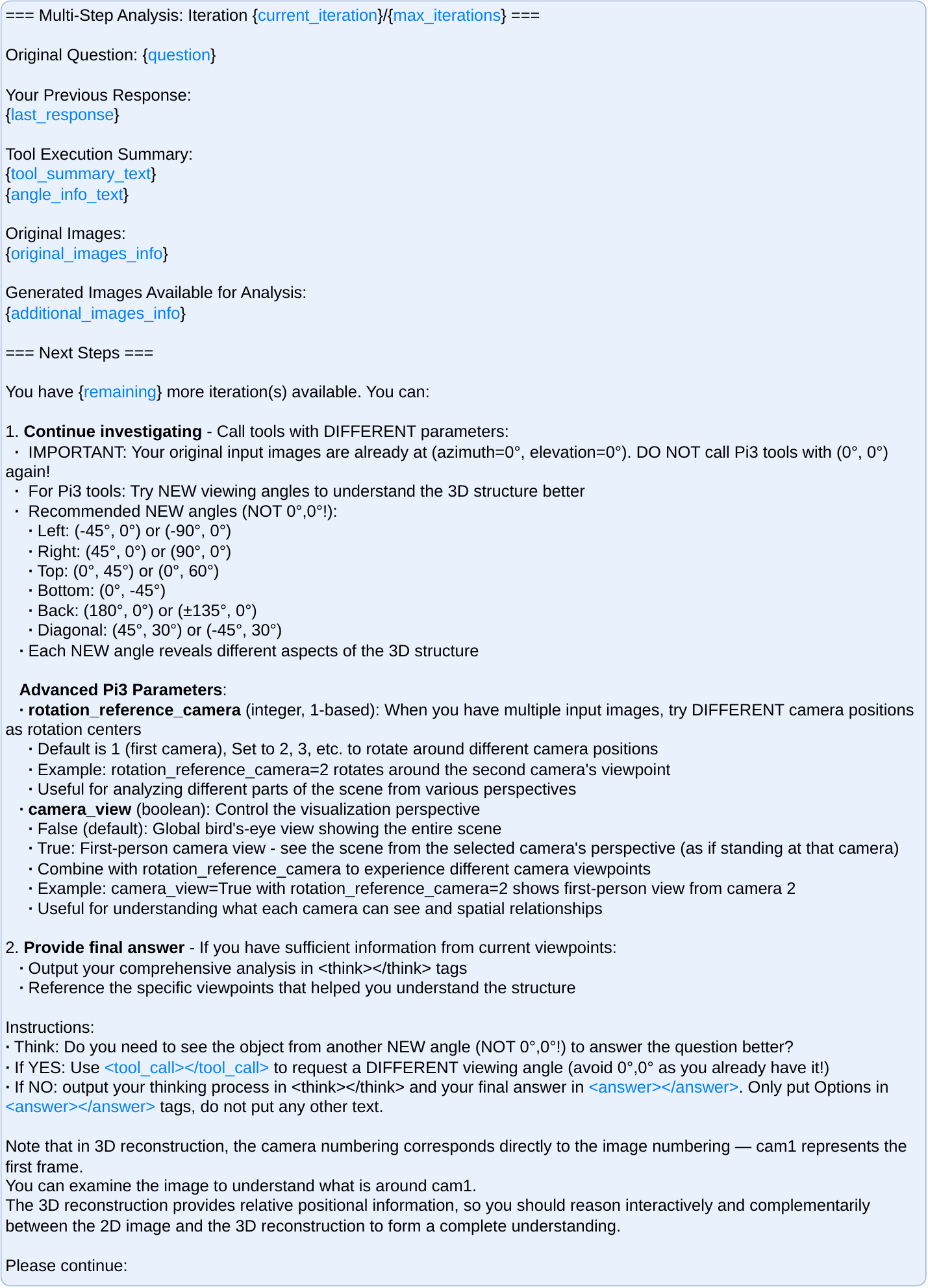}
    \caption{Multi-step prompt for iterative 3D viewpoint exploration. Including angle selection, camera rotation controls, tool invocation rules to refine spatial reasoning.}
    \Description{Continuation prompt guiding angle selection, camera rotations, tool calls, and repeated spatial reasoning.}
    \label{fig:continuation_prompt}
\end{figure}

\paragraph{Training-free Workflow Prompt.}
Following~\cite{yao2022react, li2023camel}, the prompt design in Think3D is structured into three parts: a \emph{system prompt} (Fig.~\ref{fig:system_prompt}), a \emph{tool prompt} that describes the 3D tools and their usage rules (Fig.~\ref{fig:pi3_prompt}), and a \emph{continual prompt} that updates the context at the beginning of each reasoning round (Fig.~\ref{fig:continuation_prompt}). We adopt this modular design to improve prompt clarity and make multi-round tool-augmented reasoning more stable.

The \textbf{system prompt} defines the model's role, the overall reasoning workflow, and the required output format. This helps reduce invalid tool calls and keeps the reasoning process executable and consistent. The \textbf{tool prompt} specifies how the model should use 3D reconstruction tools, including which new viewpoints are worth exploring and how to avoid redundant views. In particular, we explicitly state that the input image already corresponds to the default $(0^\circ,0^\circ)$ view, and provide recommended alternative viewpoints such as left, right, top, back, and diagonal views. This design encourages the model to request complementary observations that are more informative for spatial reasoning.

The \textbf{continual prompt} is used to maintain reasoning progress across multiple rounds. At each round, it reminds the model of the task goal, the current context, and the need to decide whether additional tool use is still necessary. This helps the model stay focused on unresolved spatial uncertainty rather than repeatedly generating redundant analysis. Overall, this three-part prompt design improves format reliability, viewpoint efficiency, and multi-round reasoning stability in the training-free workflow.

\paragraph{RL Training Prompt.}
During RL training, invoking external tools online at every rollout step is prohibitively slow, reducing training efficiency and making large-scale optimization impractical. To address this issue, we pre-generate point clouds for all training scenes in advance, thereby eliminating the need to run Pi3 inference during RL training. This design substantially reduces the per-sample processing overhead and enables more efficient policy optimization while keeping the spatial input representation consistent across training iterations.

For RL training, the system and continuation prompts specify eight
candidate angle pairs:
\begin{equation}
\begin{split}
\mathcal{V}=\{&(-45,0),(-90,0),(45,0),(90,0),\\
             &(0,60),(45,60),(45,30),(-45,30)\}.
\end{split}
\end{equation}
The selected viewpoint is rendered on demand from the precomputed point
cloud and returned to the model for the next reasoning step. Thus, scene
reconstruction is performed offline, while viewpoint selection and rendering
remain part of the interactive rollout.

In addition, we observe that smaller open-source models generally exhibit weaker instruction-following and prompt-utilization capabilities than larger proprietary models. A single fixed prompt is therefore often insufficient for stable multi-round RL optimization, especially when the model needs to progressively refine its reasoning and action prediction over repeated iterations. To improve prompt efficiency, we further divide the continual training prompts according to the current iteration stage. Specifically, different prompts are used for the initial training round, intermediate continuation rounds, and the final refinement stage. This stage-aware prompt design allows the model to receive instructions that are better aligned with its current optimization status, improving both prompt utilization and training stability.

\begin{figure}[tbp]
    \centering
    \includegraphics[width=\linewidth,height=0.88\textheight,keepaspectratio]{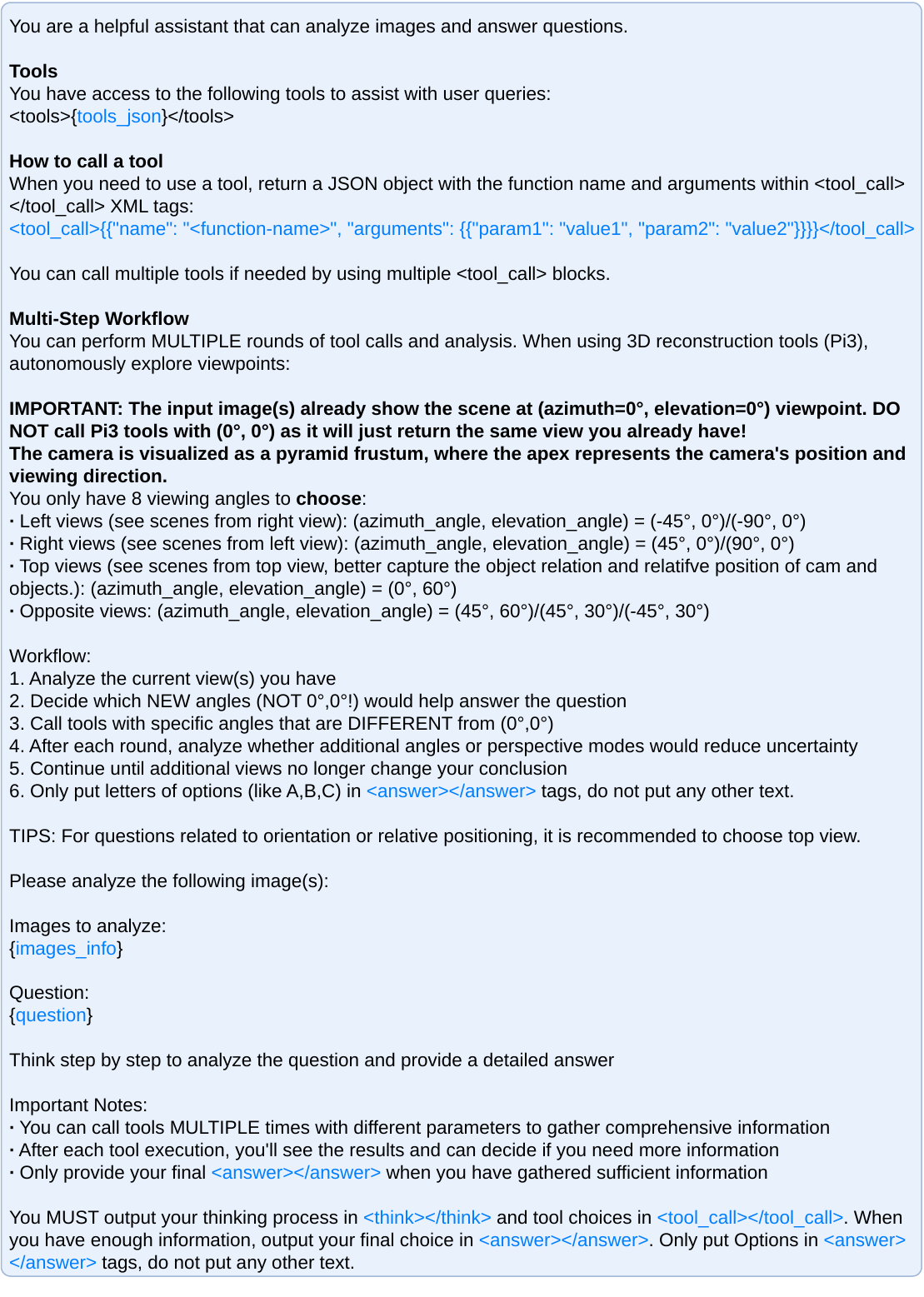}
    \caption{The RL system prompt. Instruction prompt defining the eight-candidate-view 3D analysis workflow, including tool-call format, angle selection rules (left, right, top, and oblique), and iterative reasoning steps for viewpoint-guided spatial understanding.}
    \Description{System prompt used for reinforcement learning, with an eight-candidate-view analysis workflow, tool-call format, and permitted viewpoints.}
    \label{fig:rlsystem_prompt}
\end{figure}

\begin{figure}[tbp]
    \centering
    \includegraphics[width=\linewidth,height=0.88\textheight,keepaspectratio]{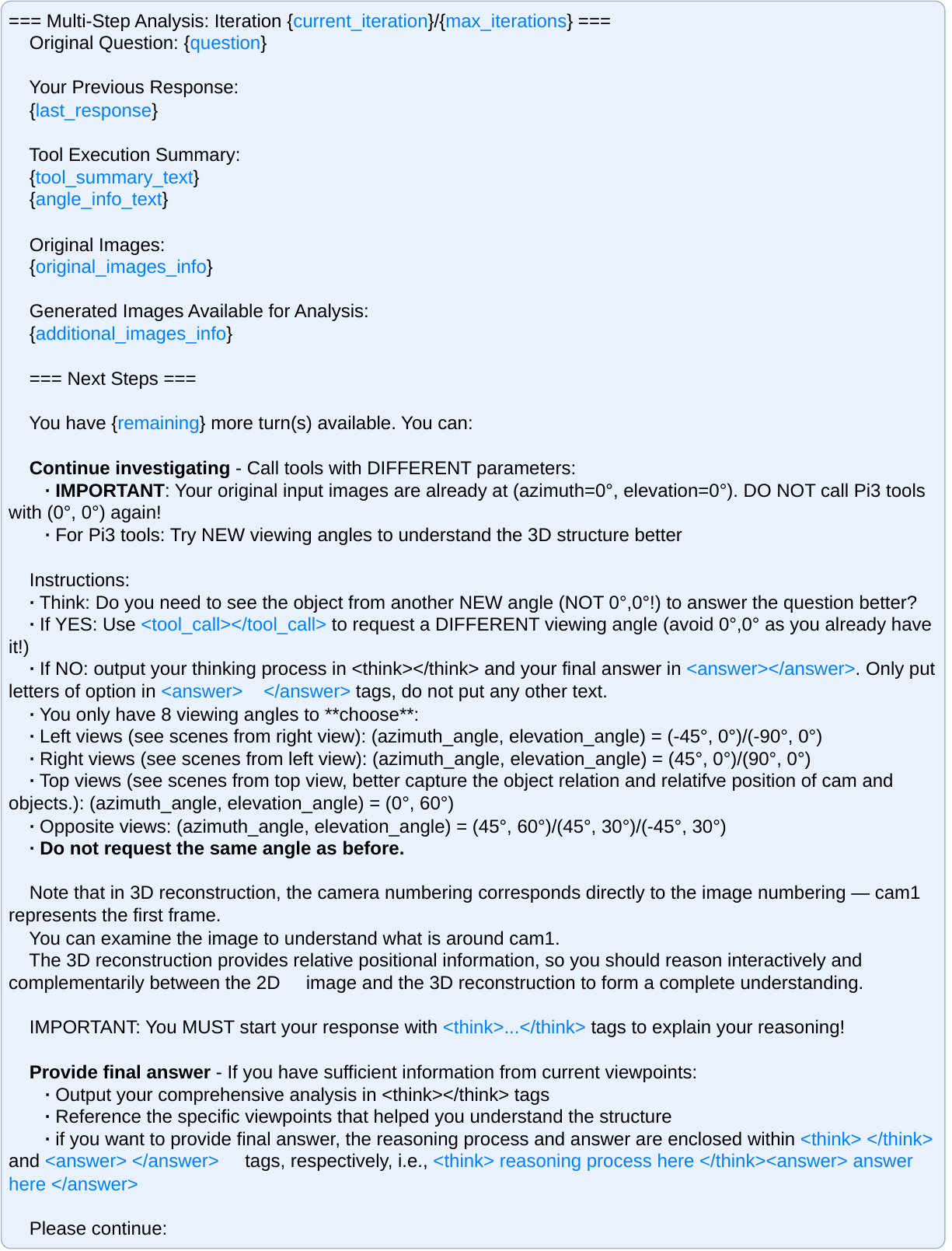}
    \caption{The RL continuation prompt used during non-final turns. Iterative-step instruction prompt outlining eight allowed viewpoint choices, tool-call rules, and the decision process for progressing or concluding 3D spatial analysis.}
    \Description{Prompt for a non-final reinforcement-learning turn, specifying allowed viewpoints and the decision to continue exploring or answer.}
    \label{fig:rlcontinue1_prompt}
\end{figure}

\begin{figure}[tbp]
    \centering
    \includegraphics[width=\linewidth,height=0.88\textheight,keepaspectratio]{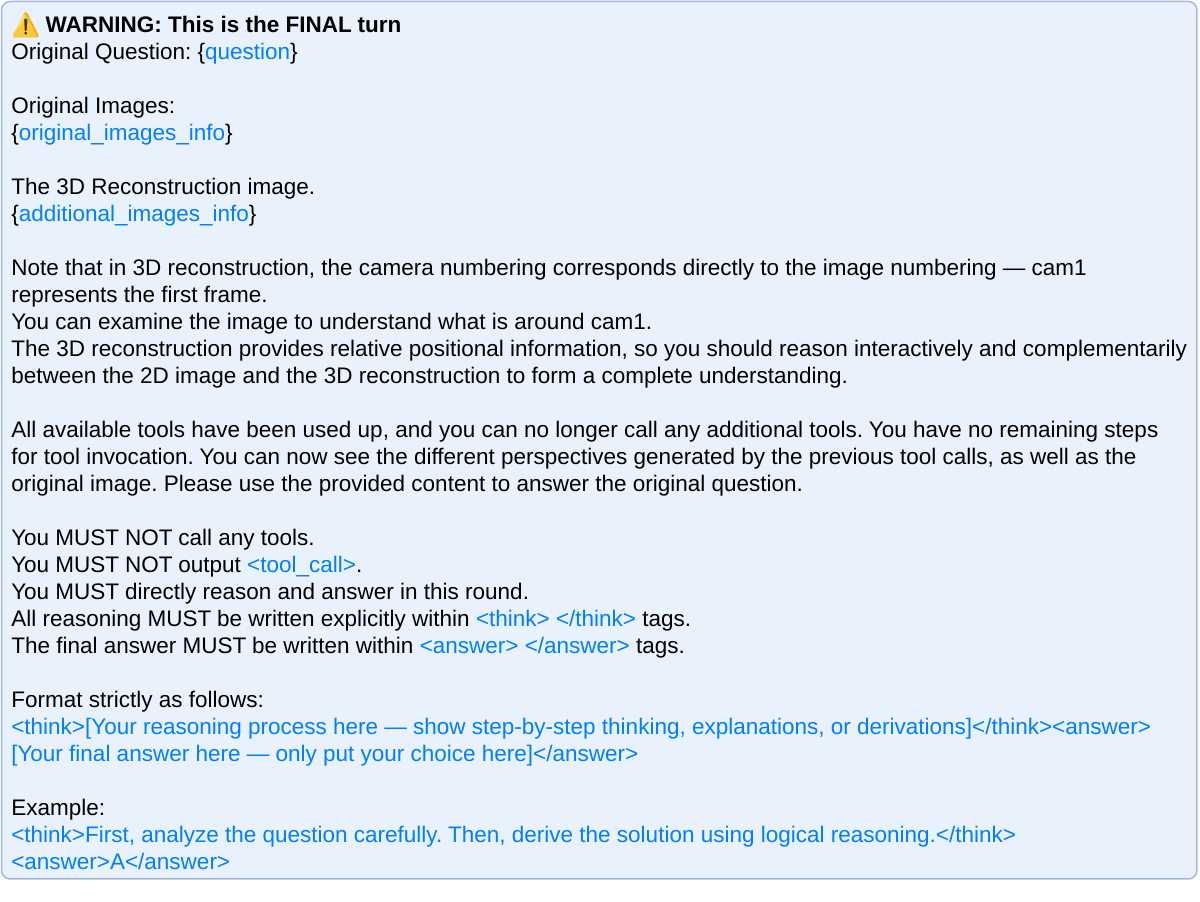}
    \caption{The RL continuation prompt used in the final turn. Final-turn instruction prompt specifying the no-tool phase, requiring explicit reasoning and a final answer based solely on previously generated 3D views and the original image.}
    \Description{Prompt for the final reinforcement-learning turn, requiring an answer based on existing views without further tool calls.}
    \label{fig:rlcontinue_final_prompt}
\end{figure}

The corresponding prompts used in different stages are shown in Fig.~\ref{fig:rlsystem_prompt}, Fig.~\ref{fig:rlcontinue1_prompt}, and Fig.~\ref{fig:rlcontinue_final_prompt}.

\begin{figure}[tbp]
    \centering
    \includegraphics[width=\linewidth,height=0.88\textheight,keepaspectratio]{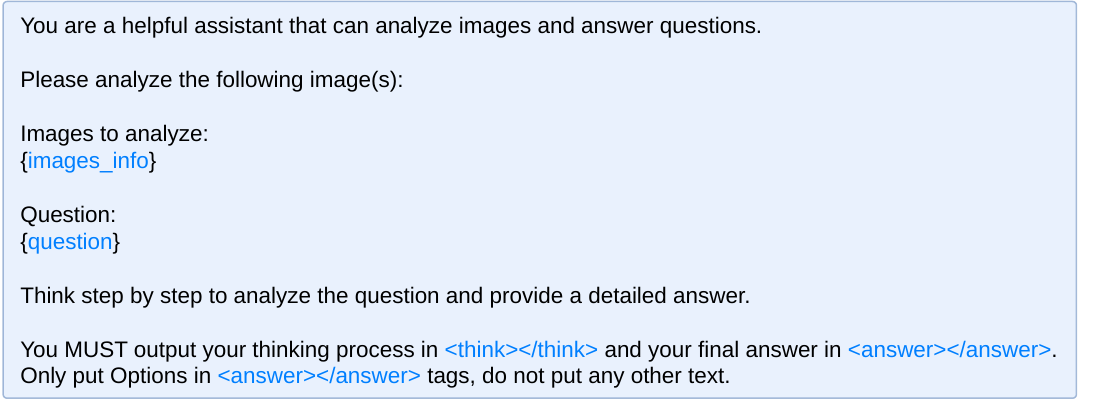}
    \caption{The prompt without tools. Base instruction prompt for direct image-question analysis, requiring explicit reasoning and final answer formatting without tool interactions.}
    \Description{Prompt for direct image-question reasoning without external tools, including reasoning and final-answer formatting.}
    \label{fig:notool_prompt}
\end{figure}

\paragraph{Prompt for evaluation without tools.} When no tool is available, we adopt standard chain-of-thought~(CoT)~\cite{wei2022chain} reasoning. The corresponding prompt is shown in Fig~\ref{fig:notool_prompt}.

\begin{figure}[tbp]
    \centering
    \includegraphics[width=\linewidth,height=0.88\textheight,keepaspectratio]{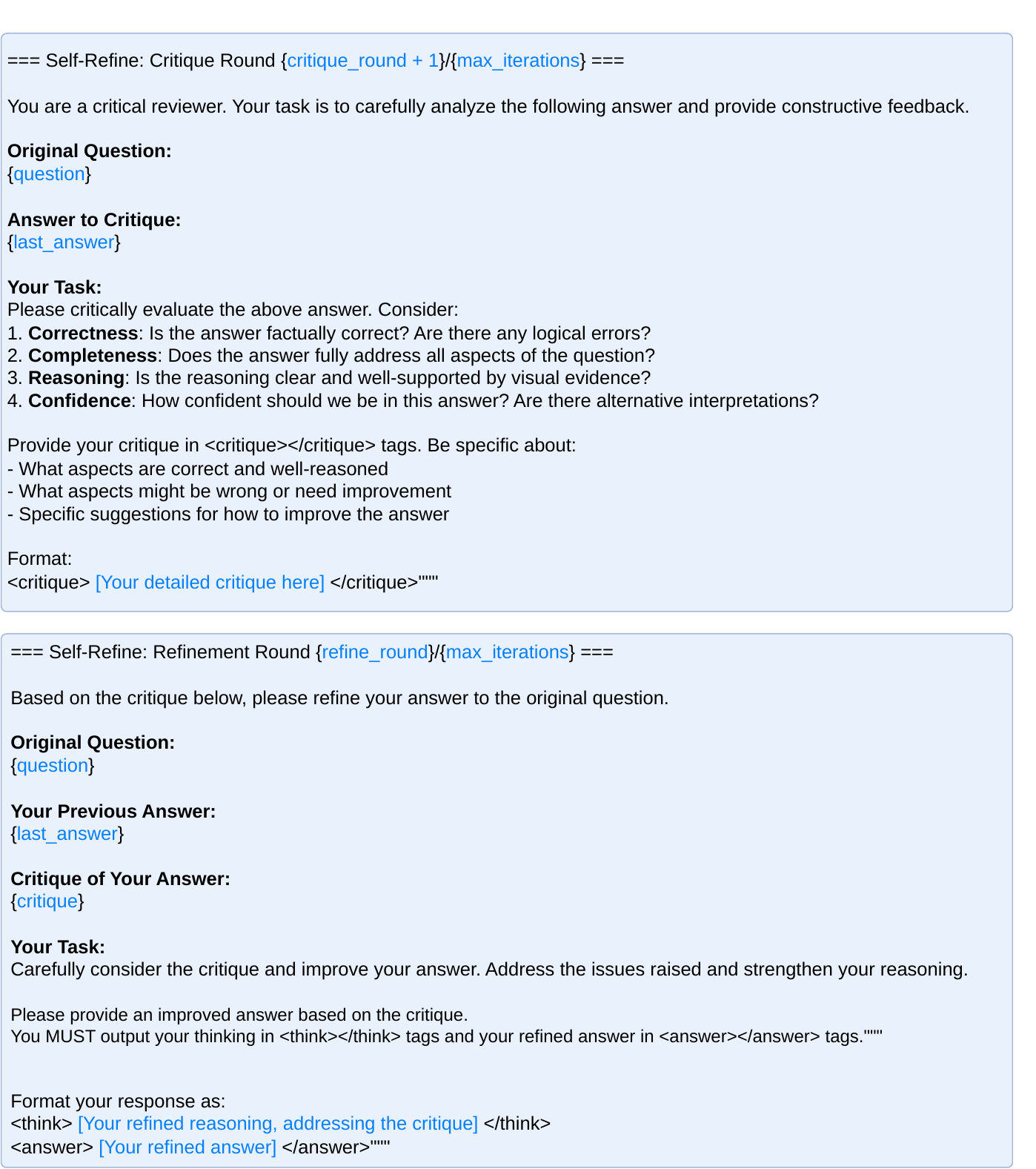}
    \caption{The critique prompt and refinement prompt used in the self-refine experiment.}
    \Description{Critique and refinement prompts used in the multi-round Self-Refine baseline.}
    \label{fig:selfrefine_prompt}
\end{figure}

\paragraph{Prompt for self-refine experiment.} The corresponding prompt is shown in Fig~\ref{fig:selfrefine_prompt}.

\begin{algorithm}[tbp]
\caption{Self-Refine Inference}
\begin{algorithmic}[1]
\REQUIRE Question $q$, iterations $T$
\STATE $a_0 \leftarrow \mathrm{LLM}(q)$
\FOR{$t=0$ to $T-1$}
\STATE $c_t \leftarrow \mathrm{LLM}_{\text{critique}}(q,a_t)$
\STATE $a_{t+1} \leftarrow \mathrm{LLM}_{\text{refine}}(q,a_t,c_t)$
\ENDFOR
\STATE \textbf{return} $a_T$
\end{algorithmic}
\end{algorithm}

\clearpage
\section{Think3D-RL Training And Evaluation Setting}
\label{app:training}
As shown in Tab~\ref{tab:rl_para}, we provide the parameters used for both RL training and evaluation. For all experiments, including the main results and ablation studies, we run each setting three times and report the average performance to ensure a fair comparison.
\begin{table}[!htbp]
\centering
\caption{Training and evaluation parameters used in both the RL optimization process and subsequent evaluation.}
\begin{tabular}{lc}
\hline
\textbf{Parameter}             & \textbf{Setting}  \\ \hline
Foundation model               & Qwen3-VL-4B-Instruct \\
Number of trained agents       & 1                 \\
Number of solution rounds      & 3                 \\
Number of evaluation rounds    & 3                 \\
Horizon for discussion history & 1                 \\
Token limit for prompts        & 180000            \\
Token limit for responses      & 1024              \\
Training temperature           & 0.6               \\
Evaluation temperature         & 1.0               \\
Clipping epsilon               & 0.2               \\
Weight of KL penalty           & 0.05              \\
Number of training epochs      & 1                 \\
Training batch size            & 32(8*4accu)       \\
Rollout batch size             & 64                \\
Optimizer name                 & AdamW             \\
Learning rate                  & 1e-6              \\
Weight decay                   & 0.1               \\
Gradient norm                  & 0.5               \\
Gradient clipping              & False             \\
Gradient checkpoint            & True              \\
Flash Attention                & True              \\
Mixed precision                & True              \\
Enable vLLM                    & False             \\
Enable DeepSpeed               & True              \\ \hline
\end{tabular}
\label{tab:rl_para}
\end{table}

\section{Results on More Models}
\label{app:models}
We provide additional evaluation results on more vision-language models in this section. The results on BLINK(Multi-view) and MindCube subset are reported in Table~\ref{tab:blink_more_models}. These results provide a broader comparison of current VLMs on spatial reasoning benchmarks.

\begin{table}[!htbp]
\centering
\caption{More results on BLINK(Multi-view) and the MindCube subset(\%).}
\scriptsize
\fitwidth{
\begin{tabular}{lccccc}
\toprule
Model & BLINK (MV) & MC (Rotation) & MC (Among) & MC (Around) & Avg \\
\midrule
InternVL3.5-1B & 51.13 & 37.50 & 35.00 & 25.00 & 36.66\\
InternVL3.5-2B & 49.62 & 37.50 & 32.50 & 42.50 & 38.46\\
InternVL3.5-4B & 53.38 & 50.00 & 30.00 & 25.00 & 36.67\\
Llava-onevision-qwen2-0.5b & 43.61 & 32.50 & 22.50 & 40.00 & 34.65\\
Llava-onevision-qwen2-7b & 45.86 & 37.50 & 25.00 & 35.00 & 35.84\\
Qwen3-VL-8B & 47.36 & 45.00 & 30.00 & 47.50 & 42.47\\
Qwen3-VL-30B-a3b-instruct & 49.62 & 37.50 & 37.50 & 52.50 & 44.28\\
GPT-4.1-mini & 49.62 & 54.17 & 39.17 & 45.83 & 47.20\\
Think3D(GPT-4.1-mini) & 52.38 & 53.33 & 47.5 & 41.67 & 48.72\\
\bottomrule
\end{tabular}
}
\label{tab:blink_more_models}
\end{table}

\FloatBarrier
\clearpage
\section{Interaction Visualization}
\label{app:visualization}
We provide additional visualization examples, as illustrated in the figures below.
\begin{figure}[H]
    \centering
    \includegraphics[width=\linewidth,height=0.78\textheight,keepaspectratio]{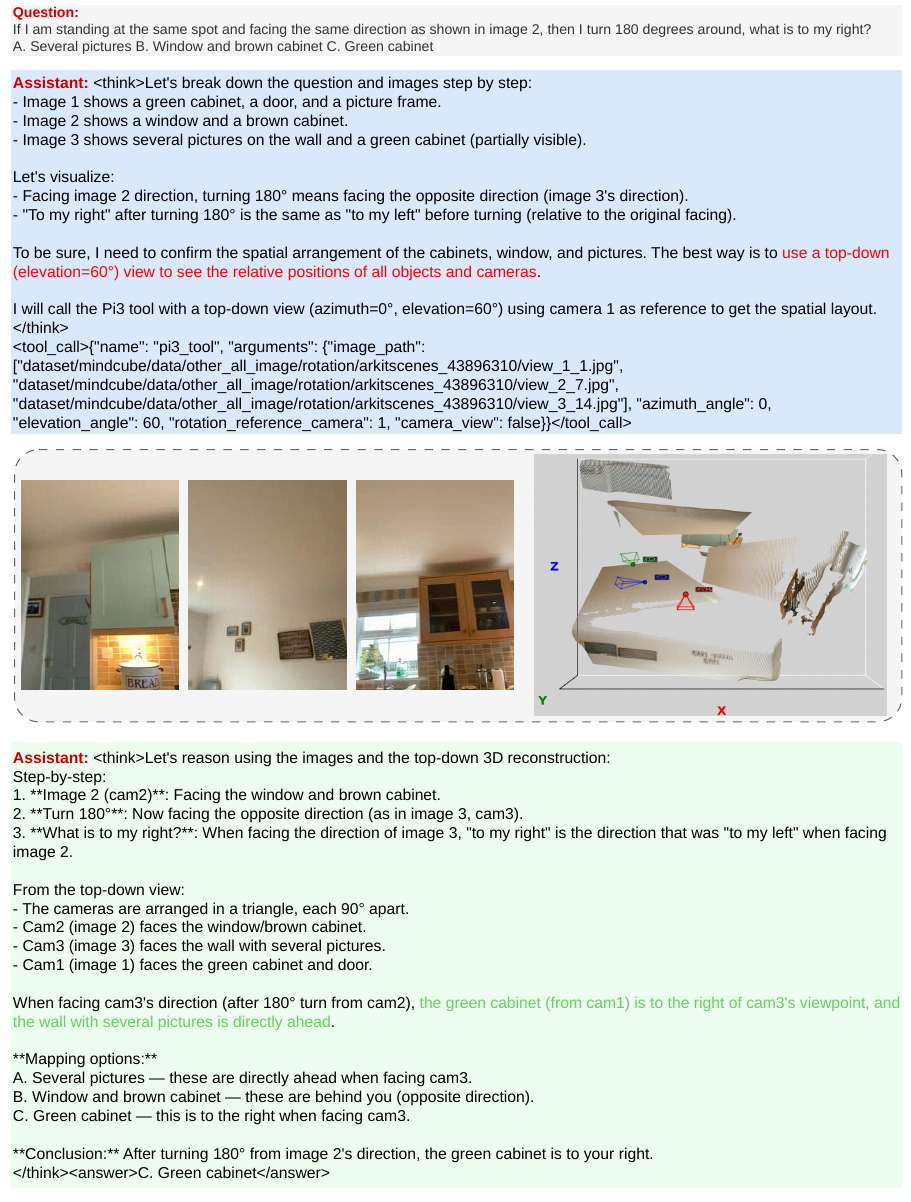}
    \caption{The Mindcube example.}
    \Description{First MindCube example showing the input, spatial exploration, and model response.}
    \label{fig:mindcube_eg1}
\end{figure}

\begin{figure*}[tbp]
    \centering
    \includegraphics[width=\linewidth,height=0.88\textheight,keepaspectratio]{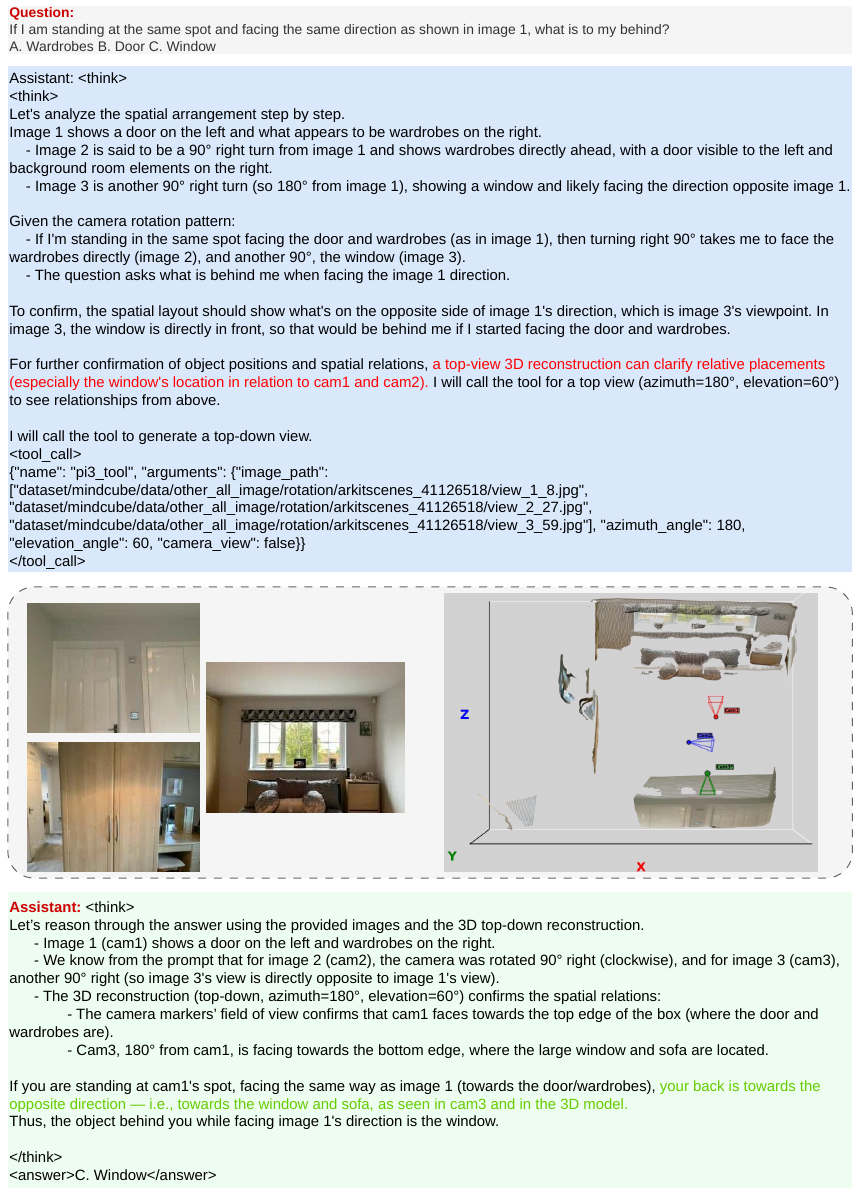}
    \caption{The Mindcube example.}
    \Description{Second MindCube example showing the input, spatial exploration, and model response.}
    \label{fig:mindcube_eg2}
\end{figure*}

\begin{figure*}[tbp]
    \centering
    \includegraphics[width=\linewidth,height=0.88\textheight,keepaspectratio]{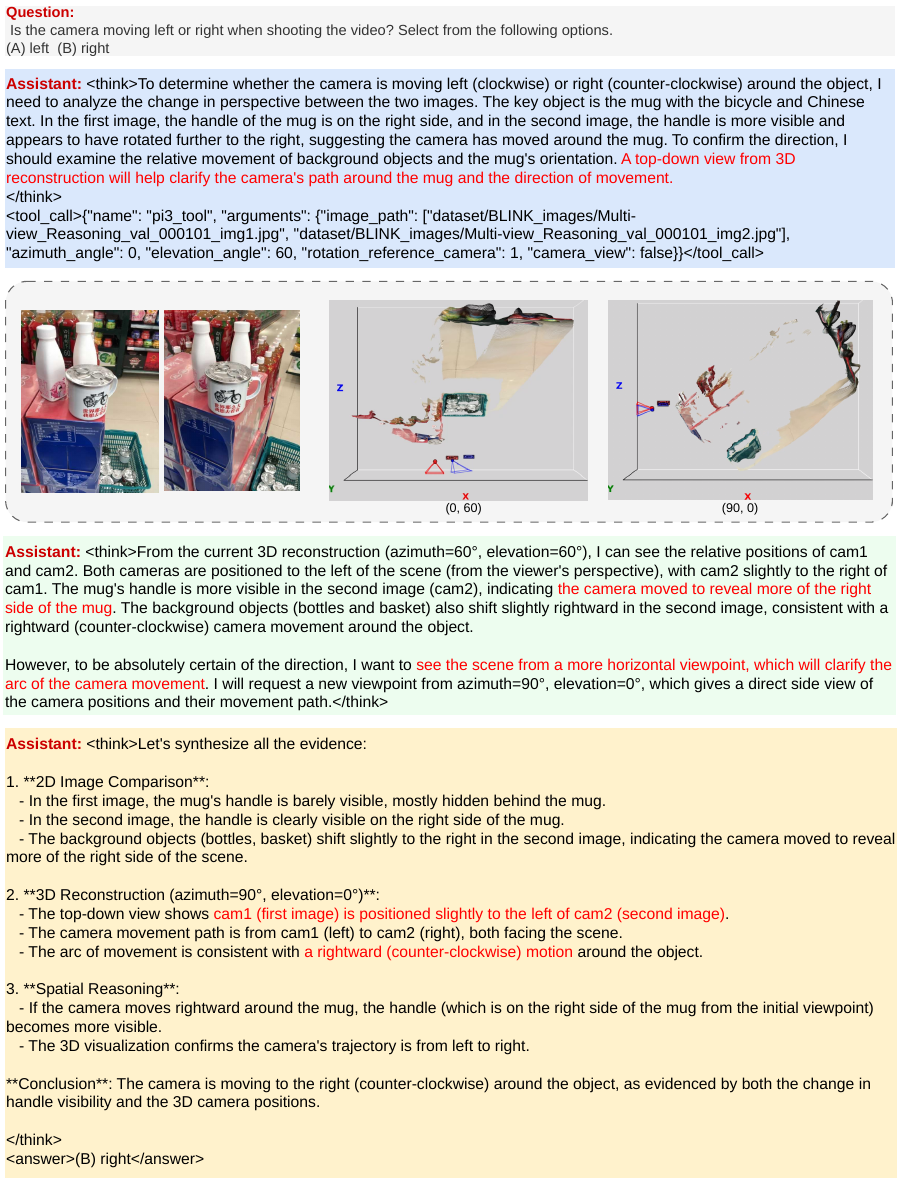}
    \caption{The BLINK example.}
    \Description{First BLINK example showing the input views, spatial exploration, and model response.}
    \label{fig:blink_eg1}
\end{figure*}

\begin{figure*}[tbp]
    \centering
    \includegraphics[width=\linewidth,height=0.88\textheight,keepaspectratio]{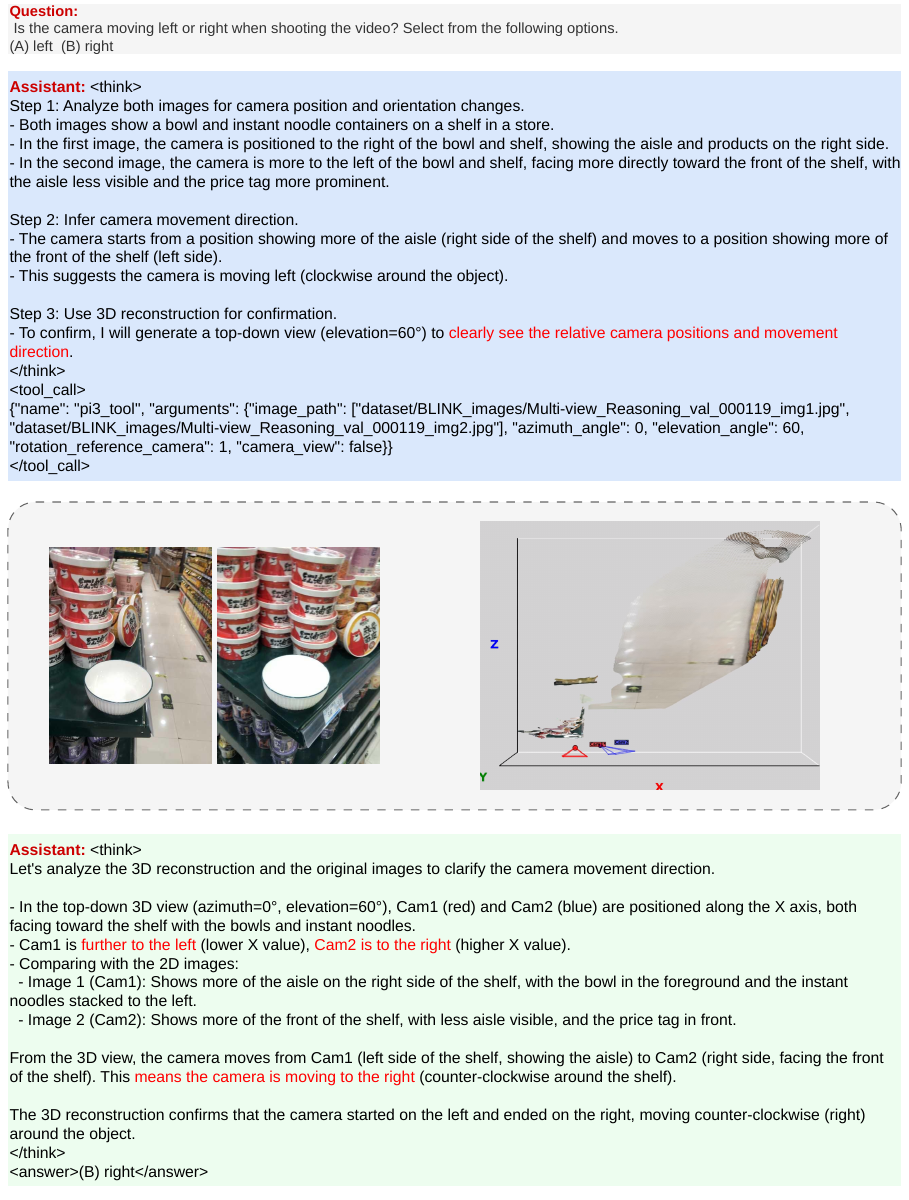}
    \caption{The BLINK example.}
    \Description{Second BLINK example showing the input views, spatial exploration, and model response.}
    \label{fig:blink_eg2}
\end{figure*}

\begin{figure*}[tbp]
    \centering
    \includegraphics[width=\linewidth,height=0.88\textheight,keepaspectratio]{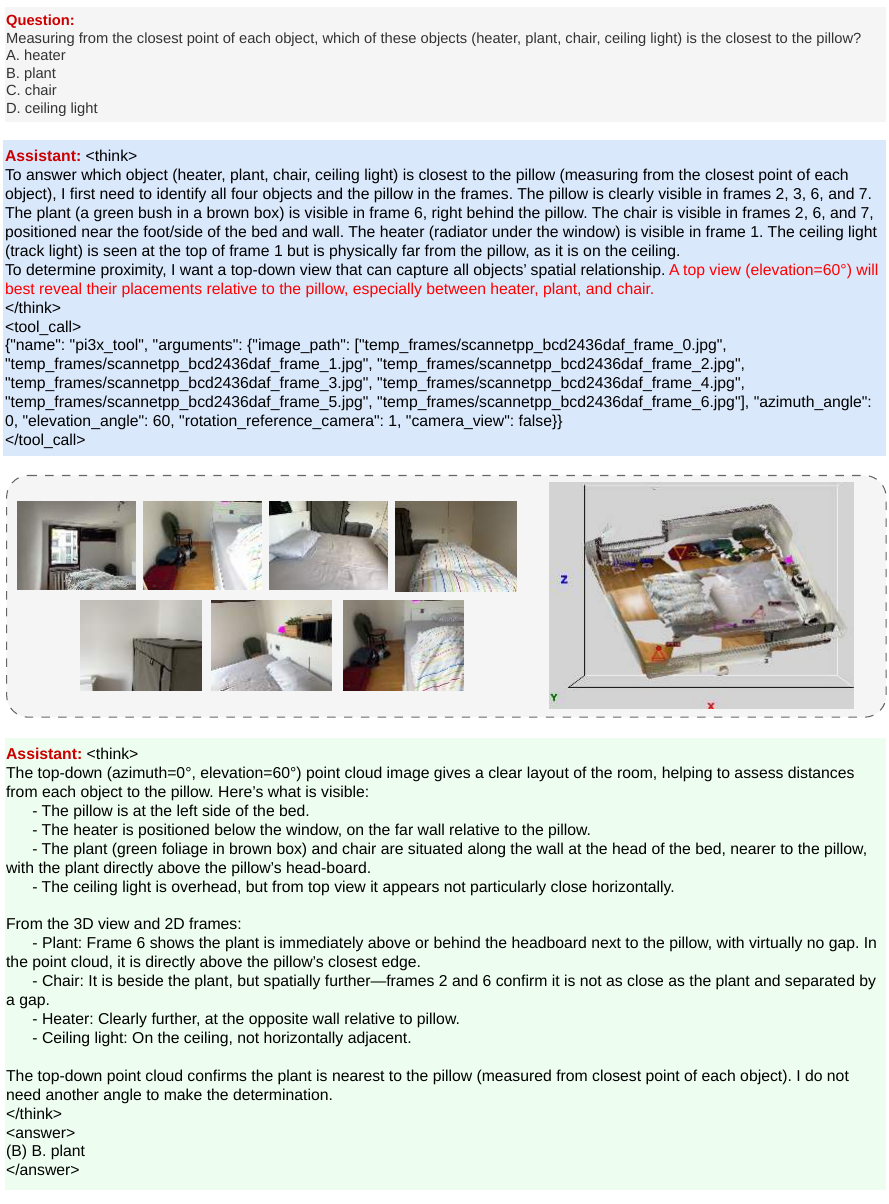}
    \caption{The VSI-Bench example.}
    \Description{First VSI-Bench example showing video frames, spatial exploration, and model response.}
    \label{fig:vsi_eg1}
\end{figure*}

\begin{figure*}[tbp]
    \centering
    \includegraphics[width=\linewidth,height=0.88\textheight,keepaspectratio]{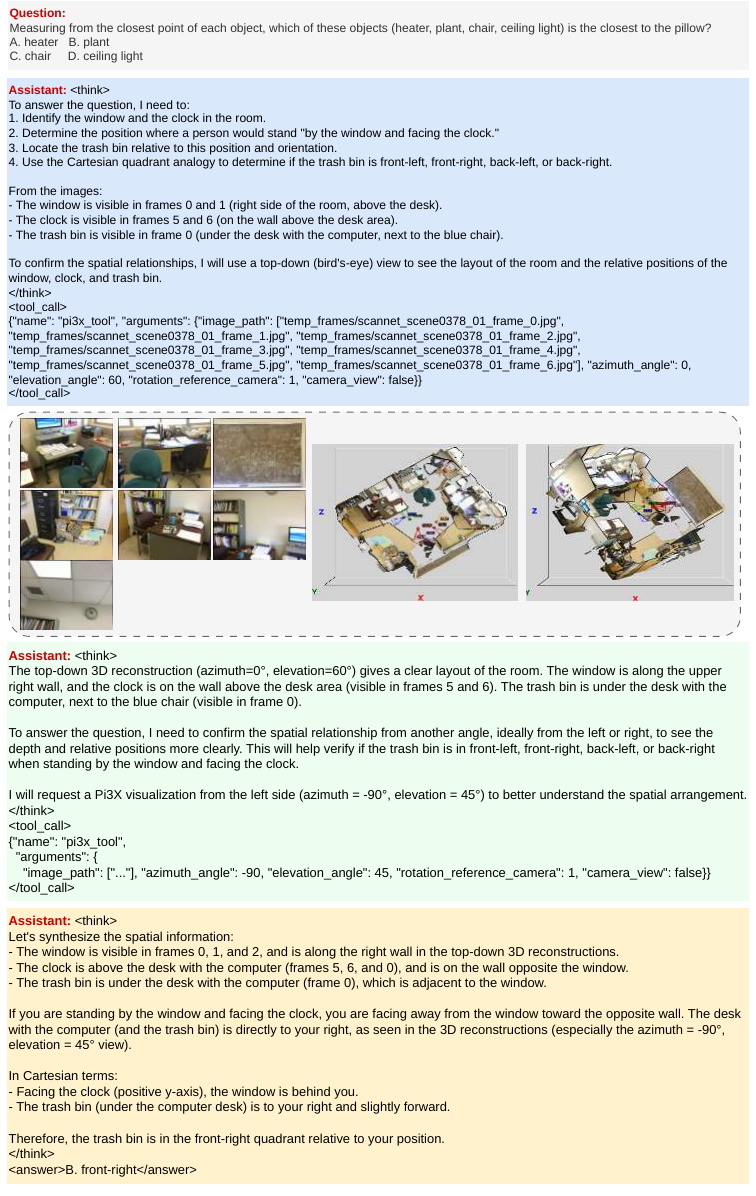}
    \caption{The VSI-Bench example.}
    \Description{Second VSI-Bench example showing video frames, spatial exploration, and model response.}
    \label{fig:vsi_eg2}
\end{figure*}

\end{document}